\documentclass[10pt,twocolumn,letterpaper]{article}

\usepackage[T1]{fontenc}
\usepackage{newtxtext}
\usepackage{helvet}
\usepackage{courier}

\usepackage[letterpaper,textwidth=7.0in,textheight=9.0in,columnsep=0.375in]{geometry}

\usepackage[hyphens]{url}
\usepackage{graphicx}
\usepackage{natbib}
\usepackage{caption}
\usepackage{booktabs}

\usepackage{array}          % table column types
\usepackage{amsfonts}       % blackboard math symbols
\usepackage{amsmath}        % math environments
\usepackage{amssymb}        % checkmark and other symbols
\usepackage{pifont}         % ding symbols for cell notation
\usepackage{xcolor}         % colors
\usepackage{placeins}       % float barrier before references

\newcommand{\method}{MentorPulse}
\newcommand{\MPname}{MP}
\newcommand{\mentor}{\mathcal{M}}
\newcommand{\student}{\mathcal{S}}
\newcommand{\mem}{Z}
\newcommand{\StaleGap}{\mathrm{StaleGap}}

\newcommand{\Bytes}{\mathrm{Bytes}}
\newcommand{\sonly}{\textsc{S-only}}
\newcommand{\monly}{\textsc{M-only}}
\newcommand{\oneshot}{\textsc{OneShot}}
\newcommand{\oracleswap}[1]{\textsc{OracleSwap}@#1}
\newcommand{\oracleswapmis}[1]{\textsc{OracleSwap-mis}@#1}
\newcommand{\mpat}[1]{\MPname @#1}
\newcommand{\specdec}{\textsc{SpecDec}}

\title{MentorPulse: Refreshing Cross-Model Latent Guidance for\\ Long-Form Generation}
\author{
  Ziwu Liu \quad Guozhong Li \quad Chen Qiu \quad Weiyang Kong \quad Panos Kalnis\\[4pt]
  King Abdullah University of Science and Technology (KAUST)
}
\date{}

\begin{document}

\maketitle

\begin{abstract}
Cross-model latent guidance lets a frozen large mentor encode an input once and a frozen small student generate from the resulting signal. Existing methods keep this signal fixed, assuming it stays useful as the output grows; we show this fails in long-form generation. On multi-turn instruction following, static guidance pushes a 4B student's constraint satisfaction 2.5 points below its no-guidance baseline; a training-free refresh every 16 tokens changes only the memory content and restores a 2.0-point gain over that baseline. We propose \method{} to keep guidance fresh at practical cost: it compresses mentor states into a capped slot memory, incrementally processes newly generated tokens, and updates the memory that the student reads through gated cross-attention without resetting the student's KV cache. Windowed Refresh Training exposes the bridge to prefix-conditioned memory. Across thirteen datasets, \method{} closes 52.2\% of the mentor--student gap on macro average, outperforming C2C, T2T, and equal-budget LoRA, with the largest gains on long outputs. It performs best on all eleven mentor--student pairs from three model families, with margins that narrow as the capability gap grows, and a lightweight read-pattern check predicts the gain before deployment. Measured costs identify refresh intervals that dominate text guidance on long outputs.
\end{abstract}

\section{Introduction}
\label{sec:intro}

\begin{figure}[t]
\centering
\includegraphics[width=\columnwidth]{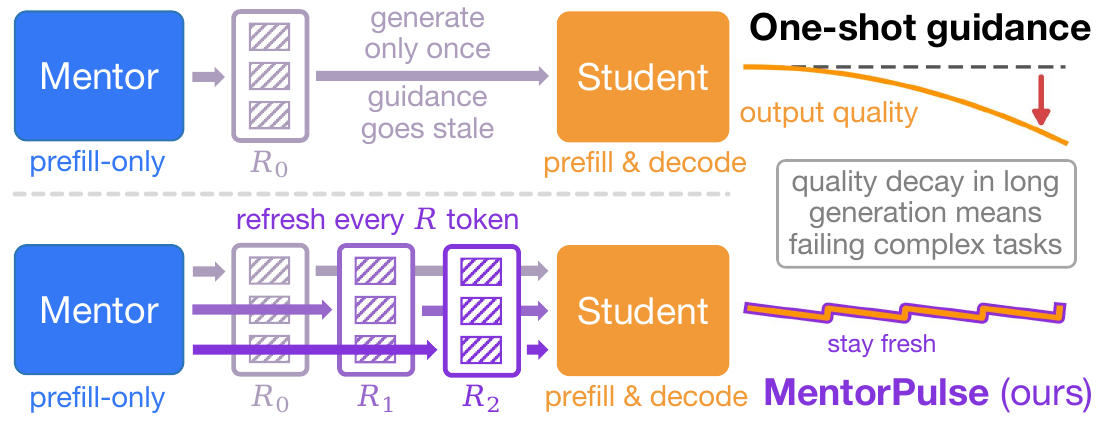}
\caption{One-shot guidance fixes the memory before decoding and quality decays over long generation (top); \method{} refreshes it from the generated prefix every $R$ tokens (bottom).}
\label{fig:concept}
\end{figure}

A frozen large \emph{mentor} can guide a frozen small \emph{student} through latent states \citep{bergner2024llm2slm,fu2026c2c,latentguided2026}. Existing methods compute this guidance before decoding and keep it fixed. In long-form generation this design fails: as the generated prefix grows, prompt-only guidance can become stale and push the student below its no-guidance baseline, while fresh, prefix-conditioned guidance reverses the degradation without changing the student (Figure~\ref{fig:concept}). We ask how to keep latent guidance fresh throughout decoding at practical cost.

% ---- Figure 1 (Introduction hook): staleness concept ----

Model capability and serving cost both grow with scale \citep{scalinglaw,B&Ssurvey1,B&Ssurvey2}. Letting a mentor process the input while a student generates retains useful large-model information at reduced decoding cost; long-form generation offers the largest saving because it has the most decoding steps, yet there fixed guidance is most likely to become outdated. A compact, refreshable memory further extends prefill--decode disaggregation \citep{zhong2024distserve,patel2024splitwise} across model sizes and networked devices.

Prior methods establish that a student can use mentor states through a learned interface: LLM-to-SLM transfers a one-time prompt encoding \citep{bergner2024llm2slm}, C2C fuses the mentor's KV cache into the student's cache \citep{fu2026c2c}, and Latent-Guided Reasoning transfers compact strategy vectors \citep{latentguided2026}. All keep the signal fixed during student decoding---usually harmless for short outputs, but untested systematically for long outputs, where the generated prefix changes the relevant context.

Section~\ref{sec:diagnosis} isolates freshness with a training-free intervention: at fixed intervals the mentor prefills the input and generated prefix, then replaces the memory, leaving the student's weights and KV cache unchanged. On affected long-output tasks, static guidance falls below the student-only baseline and fresh guidance restores a gain; fresh but mismatched memory does not help, attributing the repair to updated content. Tasks beyond the student's capability do not improve, and short outputs finish before the first update. The diagnostic thus separates staleness from capability limits, yet reprocessing the growing prefix at every update makes it too expensive to deploy.

\method{} makes this repair efficient. It compresses mentor states into a capped, position-aware slot memory that the frozen student reads through gated cross-attention. Versioned incremental refresh reuses the mentor's KV cache to process only new tokens, then replaces the memory without invalidating the student's decoding history; Windowed Refresh Training optimizes the output window of each sampled memory version. The interval $R$ controls the quality--cost trade-off, and $R\to\infty$ recovers one-shot guidance.

Across thirteen datasets, \method{} closes 52.2\% of the mentor--student gap on the main pair, about twice the recovery of the strongest prior alternative. Its advantage is concentrated on long outputs, where static guidance becomes harmful, and it performs best on all eleven pairs from three model families; the margin narrows on the most capability-separated pairs, tracked in advance by our read-distribution indicator $V_{64}$. Measured throughput places the default $R{=}16$ among the cost-efficient settings $\{8,16,32\}$ against T2T on long-output tasks.

Our contributions are: (1) we identify guidance staleness as a failure mode of one-shot guidance and separate it from student capability limits via a training-free diagnostic; (2) we introduce \method{}, combining versioned slot memory, incremental mentor prefill, and Windowed Refresh Training without changing the student's decoding cache; (3) we demonstrate generality across eleven pairs and three model families, and propose the 64-step read-distribution variance $V_{64}$ to screen pairs before deployment; (4) we provide byte-level refresh accounting and measure when \method{} is more cost-efficient than text guidance on long-output tasks. \method{} thus turns long-form generation from the regime where one-shot guidance fails into the one where latent guidance pays off most.

\section{Related Work}
\label{sec:related}

\noindent
\textbf{One-shot cross-model guidance.}
The closest predecessors transfer input-conditioned latent state across models. LLM-to-SLM encodes the prompt once with a frozen large model and conditions a smaller model on the projected representations \citep{bergner2024llm2slm}. C2C projects and fuses a Sharer's input KV cache with a Receiver's cache layer by layer \citep{fu2026c2c}, and Latent-Guided Reasoning compresses a large model's solution strategy into a fixed set of latent vectors for a smaller reasoner \citep{latentguided2026}; related single-model work compresses prompts or context into latent summaries \citep{softprompt,gist,icae,autocompressors}. The transferred objects differ, but the schedule is shared: the signal is set before the receiving model decodes and is never revised from its evolving output. \method{} takes the step LLM-to-SLM names as a future direction, refreshing the transferred signal from the generated prefix while preserving the student's autoregressive state; among these interfaces, only \method{} combines a frozen mentor, an explicit state-count cap, and prefix-conditioned refresh (Table~\ref{tab:paradigms}, appendix).

\noindent
\textbf{Other large--small model collaboration.}
Distillation transfers teacher behavior into student parameters, and LoRA adapts a restricted weight subset; both yield persistent parameters rather than input-specific latent state \citep{hinton2015distilling,kd1,hu2022lora,cotb2s1}. Routing and cascading conditionally invoke one or several models but exchange selections or text, not hidden state consumed by a continuously decoding student \citep{router1,router2,cascading,B&Ssurvey1,B&Ssurvey2}. Speculative decoding is lossless acceleration: the target model verifies cheap drafts and preserves its own distribution, at the price of engaging both models at every step \citep{leviathan2023fast,chen2023accelerating,eagle,speculativesurvey}. \method{} instead leaves generation to the student while the mentor supplies intermittent prefix-conditioned state.

\noindent
\textbf{Dynamic state within one model.}
Prior work also updates state during inference, but produces and consumes it within one model: RefreshKV rebuilds a partial KV cache with occasional full-attention steps during long generation \citep{xu2024refreshkv}, and MemoryLLM updates a fixed-size latent memory from new text \citep{wang2024memoryllm}. These address attention efficiency and knowledge updating; \method{} refreshes guidance produced by a separate mentor while leaving the student's own KV cache untouched.

\noindent
\textbf{Disaggregated serving.}
Prefill--decode systems place the two inference phases of one model on different workers and ship request-specific KV state between them \citep{zhong2024distserve,patel2024splitwise,mooncake}; the unit of disaggregation is a phase of one model, not semantic guidance between different models. \method{} instead uses a boundary across model scales: a mentor produces semantic guidance that a separate student consumes, over a capped, prefix-independent payload. All these lines leave open the behavioral question central to this paper---can fixed input-only guidance become stale enough to harm the student as its generated prefix grows?---which Section~\ref{sec:diagnosis} tests directly.

\section{Guidance Staleness in Long-Form Generation}
\label{sec:diagnosis}

\subsection{Setup}
\label{sec:diag-setup}

To test whether input-only guidance stays useful throughout long-form decoding, we pair a frozen Qwen3.5-27B mentor with a frozen Qwen3.5-4B student (A2) \citep{qwen35blog}. The mentor prefills the input $x$ once, its states are compressed into a slot memory $\mem$, and the student reads $\mem$ at every layer through a gated cross-attention bridge. Only the bridge is trained; with all gates at zero the system reduces to the plain student. We compare \oneshot{}, the static bridge, against \sonly{}, the student alone as a floor, and \monly{}, the mentor alone as a ceiling. Five diagnostic tasks, disjoint from the Section~\ref{sec:experiments} benchmarks and bridge training data, are chosen for expected staleness sensitivity: Multi-IF \citep{he2024multiif}, QMSum \citep{zhong2021qmsum}, and HelloBench long-form writing \citep{que2024hellobench} as positives, BigCodeBench \citep{zhuo2025bigcodebench} and ARC-Challenge \citep{clark2018arc} as controls. Prior work established that mentor states carry information the student cannot form alone \citep{bergner2024llm2slm,fu2026c2c,latentguided2026}; we ask how long it stays valid.

% ---- Figure: oracle diagnosis, behavior (a) and representation (b) ----
\begin{figure}[t]
\centering
\includegraphics[width=\columnwidth]{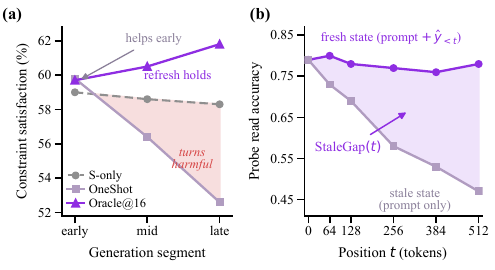}
\caption{Staleness diagnosis on Multi-IF. (a) Effect: constraint satisfaction (fraction of verifiable constraints satisfied) by generation segment. (b) Mechanism: probe read accuracy for upcoming-window properties; the shaded wedge is $\StaleGap(t)$.}
\label{fig:oracle}
\end{figure}

\subsection{Staleness and the Oracle}
\label{sec:oracle}

One-shot guidance is computed from $x$ at $t{=}0$, while the decision at step $t$ is conditioned on $x \oplus \hat{y}_{<t}$. To quantify this mismatch, let $q_t$ denote verifiable properties of a fixed-length window starting at position $t$, and let $A_{\mathrm{stale}}(t)$ and $A_{\mathrm{fresh}}(t)$ be the accuracies of matched linear probes reading $q_t$ from $H_{\mentor}(x)$ and $H_{\mentor}(x\oplus\hat{y}_{<t})$. We define $\StaleGap(t)=A_{\mathrm{fresh}}(t)-A_{\mathrm{stale}}(t)$. The probes act on mentor states directly, without the bridge (Appendix~\ref{app:probe}). On Multi-IF, $A_{\mathrm{stale}}$ falls from 0.79 at $t{=}0$ to 0.47 at $t{=}512$, while $A_{\mathrm{fresh}}$ stays between 0.76 and 0.80 (Figure~\ref{fig:oracle}b): the guidance drifts away from the task it describes.

Representation drift does not itself prove behavioral harm, so we intervene with a condition changing only freshness. \oracleswap{16} pauses decoding every 16 generated tokens, re-prefills $x \oplus \hat{y}_{<t}$ with the mentor, and swaps the rebuilt memory in place; the student's weights, generated text, and KV cache are untouched, and nothing is trained. \emph{Oracle} means full recomputation, not access to labels or future tokens. On Multi-IF, \oneshot{} scores 56.2, below the \sonly{} floor of 58.7 by more than twice the pooled standard deviation: static guidance actively misleads a student trained to trust the memory. \oracleswap{16} lifts the score to 60.7, 4.5 above \oneshot{} and 2.0 above the floor; QMSum and HelloBench move the same way (+2.7 and +3.1 over \oneshot{}). \oneshot{} helps early, crosses below the floor mid-generation, and collapses late, while \oracleswap{16} holds the late segment (Figure~\ref{fig:oracle}a). Because the swap changes nothing but the memory content, the repair can only come from freshness; equally fresh memory from a different sample brings no repair (Appendix~\ref{app:conditions}). The effect has boundaries: on BigCodeBench all conditions sit within noise, a capability failure that refresh cannot fix, and on ARC-Challenge outputs end before the first swap, so \oracleswap{16} equals \oneshot{} exactly; Section~\ref{sec:analysis} maps these boundaries across pairs.

The oracle repairs the failure but cannot be deployed: every swap re-prefills the entire growing prefix, so the cost of staying fresh grows with generation. A practical mechanism must process only new tokens, deliver updates without invalidating the student's decoding history, and train the bridge for mid-generation memory changes; Section~\ref{sec:method} meets all three.

\section{\method{}}
\label{sec:method}

\begin{figure*}[t]
\centering
\includegraphics[width=\textwidth]{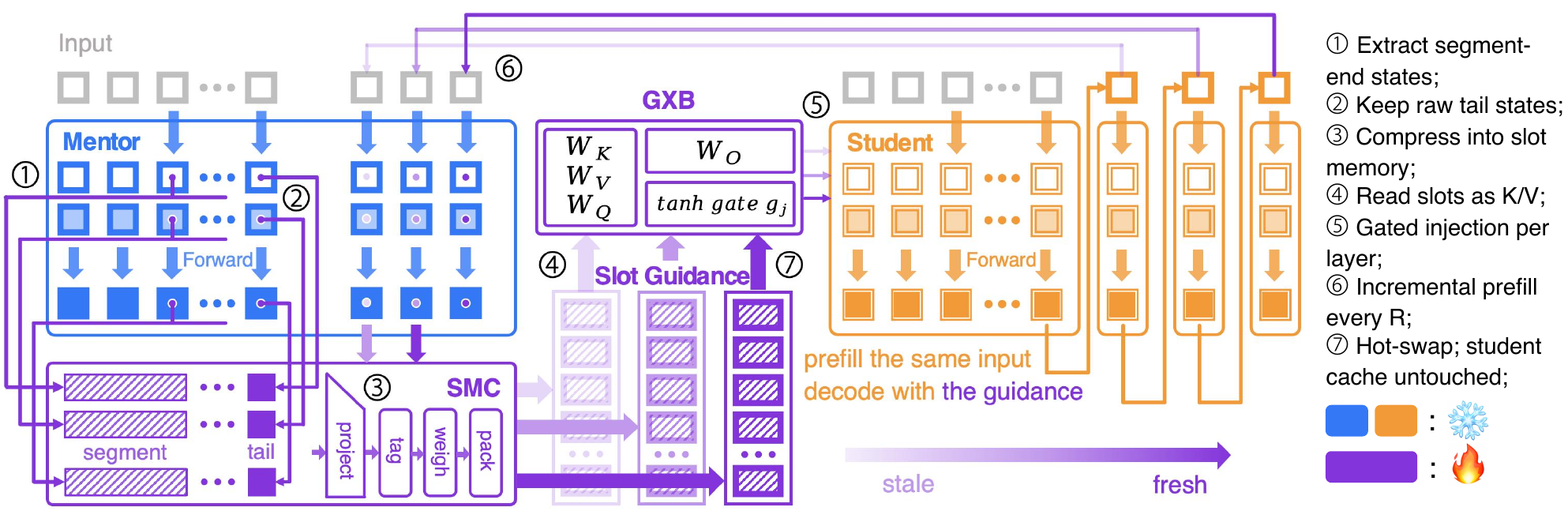}
\caption{\method{} data flow. The frozen mentor (blue) prefills, the SMC compresses its states into slot guidance, and the frozen student (orange) reads the slots through the gated bridge; every $R$ tokens an incremental prefill refreshes the memory in place (steps \ding{172}--\ding{178} in the text). Purple components are trained; saturation encodes freshness.}
\label{fig:arch}
\end{figure*}

The oracle of Section~\ref{sec:oracle} shows what freshness is worth; \method{} delivers it under deployment constraints. Five commitments shape the design: both backbones stay frozen and only a pair-specific bridge is trained; the mentor prefills and never decodes; guidance lives in a slot memory separate from the student's KV cache, so an update replaces one tensor; the slot count is capped, giving every transfer a closed-form byte count; and a single interval $R$ sets how often guidance is renewed. Figure~\ref{fig:arch} shows the flow.

% ---- Figure: system data flow (double column) ----

\subsection{Slot Memory Constructor}
\label{sec:smc}

At $t{=}0$ both models prefill the input $x$, and the Slot Memory Constructor (SMC) turns the mentor's residual-stream states into slots (\ding{172}--\ding{174}). The newest $W_{\mathrm{tail}}{=}32$ positions are kept as raw states, preserving exact wording near the context boundary. Earlier positions are split into $S{=}32$-token segments, each contributing the true state at its final position rather than a pooled average (Table~\ref{tab:ablation}, row 9); at most $P{=}128$ prompt segments are kept, longer inputs getting evenly coarser segments. For each mentor layer $o$, states are normalized by a fixed scalar $\mathrm{rms}_o$ and projected from the mentor width $d_m$ to the memory width $d_t$ (equal to the student hidden size $d_s$) by a shared map with a rank-64 per-layer correction; embeddings tag mentor layer, slot order, and segment type, and a learned weight $\rho_o$ scales each layer's slots; sorting layers by $\rho_o$ sets the transmission pruning order (row 11). With $k$ transmitted layers and $p_x\leq P$ prompt segments, the initial memory is $\mem_0\in\mathbb{R}^{k(p_x+W_{\mathrm{tail}})\times d_t}$. Two properties matter downstream: the capped slot count keeps a transfer from growing with input length as a full KV cache would, and position-based segmentation needs no cross-model token alignment.

\subsection{Gated Cross-Attention Bridge}
\label{sec:gxb}

The student reads the current memory after each layer $j$ through gated cross-attention (\ding{175}--\ding{176}), $h_j\leftarrow h_j+\tanh(g_j)\,\mathrm{CrossAttn}(Q{=}h_j,\;K,V{=}\mem_t)$: the query is the student residual stream, the memory supplies keys and values, and all four attention projections are low rank ($r{=}256$). The gates start at zero, so the untrained bridge is an identity map and the system reproduces the plain student bit for bit---the degeneracy check of Sections~\ref{sec:diagnosis} and~\ref{sec:ablation}. Training escapes this dead zone by warm-starting from a static-bridge checkpoint whose gates are already open (Appendix~\ref{app:method}). Because $\mem_t$ is a side tensor outside the student's self-attention cache, replacing it never invalidates cached states---the basis of refresh.

\subsection{Versioned Incremental Refresh}
\label{sec:refresh}

Every $R$ generated tokens (\ding{177}), the student sends its newest tokens to the mentor, which extends its own KV cache and computes states only for the new positions; by causality, earlier states are unchanged. Across a whole generation each token is therefore prefilled exactly once, and the mentor's total compute is independent of $R$, a fact the cost analysis of Section~\ref{sec:cost} builds on. One caveat: new queries still attend to the cached prefix, so a single refresh's attention cost can grow with accumulated context length.

The memory updates incrementally: the span covered by the previous tail is re-segmented into generated-prefix summary slots, a fresh tail is appended, and existing summary slots are never resent. The segment-type embedding distinguishes prompt summaries, generated-prefix summaries, and the current tail (row 12). The amortized bf16 payload per update is
\begin{equation}
\Bytes(R)=k\left(R/S+W_{\mathrm{tail}}\right)d_t\,b,
\label{eq:bytes}
\end{equation}
with $b$ bytes per value: a new summary slot completes only every $\lceil S/R\rceil$-th refresh, so $R/S$ is the average summary payload per update, and the worst single update ships $\lceil R/S\rceil$ summary slots. For fixed $R$, $k$, $S$, and precision, the payload is independent of the accumulated prefix length and dominated by the tail slots. For A2 at $R{=}16$ the update is 10.6\,MB with all 64 mentor layers ($k{=}64$) and 2.7\,MB with the top 16 ($k{=}16$, row 11). At the $j$-th refresh the student swaps $\mem_{t_{j-1}}$ for $\mem_{t_j}$ between two decoding steps (\ding{178}); its weights, generated text, and self-attention cache stay untouched. $R\to\infty$ recovers the static bridge.

\subsection{Windowed Refresh Training}
\label{sec:wrt}

A bridge trained only on fixed memory has never seen its guidance change; Section~\ref{sec:ablation} (row 3) shows what that costs. Windowed Refresh Training (WRT) closes the gap while preserving parallel teacher forcing. For a pair $(x,y)$, one mentor pass over $x\oplus y$ contains, by causality, the states of every refresh version. Each training row samples a boundary $t\in\{0,R,2R,\ldots\}$, builds $\mem_t$ from the prompt and the label prefix $y_{\leq t}$, and applies cross-entropy, over the bridge parameters $\theta$ only, to the window $(t,t{+}R]$, the span that decodes under $\mem_t$ at deployment:
\begin{equation}
\mathcal{L}_{\mathrm{WRT}}=-\mathbb{E}_{(x,y),t}
\sum_{i=t+1}^{\min(t+R,|y|)}
\log p_\theta(y_i\mid x,y_{<i},\mem_t).
\label{eq:wrt}
\end{equation}

Two approximations remain: prefix positions read the single latest $\mem_t$ although deployment decoded them under older versions, and label prefixes are a train--test mismatch because inference conditions on student-generated prefixes (Appendix~\ref{app:training}). Sampling $t{=}0$ keeps static training in the objective, and WRT warm-starts from the static checkpoint with new embeddings zero-initialized. \method{} needs no mentor decoding at inference; labels are generated offline.

\section{Experimental Evaluation}
\label{sec:experiments}

We ask whether refreshed latent guidance outperforms the student-only, static-latent, text-guidance, and parameter-adaptation baselines across input and output lengths; which components produce the gains and whether they generalize across mentor--student pairs; and how the refresh interval trades off quality, communication, and serving cost.

\subsection{Experimental Setup}
\label{sec:setup-exp}

\paragraph{Hardware settings.}
The serving topology matches the deployment that Section~\ref{sec:cost} prices: the main-pair mentor runs on one data-center GPU (NVIDIA H100 80GB) and the student on one consumer GPU (NVIDIA RTX 4090 24GB), connected cross-site at an effective 150\,Mbps with 18\,ms round-trip latency. Both models are served with vLLM 0.19.1 in bf16; every throughput entering the cost model is measured on this stack at batch size 1 with an 8K-token context (measured rates, rental prices with quotation dates, and network and billing sensitivity analyses in Appendix~\ref{app:bytes}). All compared methods use the same serving stack and decoding configuration within each pair; the one registered exception is the Gemma-4 pair, served through HuggingFace transformers rather than vLLM for every condition alike (Appendix~\ref{app:env}).

\paragraph{Models and benchmarks.}
We evaluate eleven mentor--student pairs from three model families (Qwen, Gemma-4, and Ministral 3), with 1.7B--12B students; pairs sharing a mentor form a \emph{mentor group} (A--E), and the main pair A2 is Qwen3.5-27B $\to$ Qwen3.5-4B. Thirteen datasets cover short-input selection (A--C: MMLU-Pro \citep{wang2024mmlupro}, GPQA \citep{rein2024gpqa}, AGIEval-MCQ \citep{zhong2024agieval}), short-input generation (D--H: MATH-500 \citep{hendrycks2021math,lightman2024verify}, OlympiadBench \citep{he2024olympiadbench}, LiveCodeBench \citep{jain2025livecodebench}, IFEval \citep{zhou2023ifeval}, WritingBench \citep{wu2025writingbench}), long-input selection (I--J: LongBench~v2 \citep{bai2025longbenchv2}, QuALITY \citep{pang2022quality}), and long-input long-output generation (K--M: GovReport \citep{huang2021govreport}, MultiNews \citep{fabbri2019multinews}, LongBench-Write \citep{bai2025longwriter}).

\paragraph{Compared methods.}
\sonly{} and \monly{}, the student and mentor alone, define the two anchors. T2T prepends a short mentor-generated guidance text to the student's prompt; C2C fuses the mentor's layer-wise KV cache into the student's cache once before decoding \citep{fu2026c2c}; LoRA fine-tunes the student with the same trainable parameter count as the \method{} bridge \citep{hu2022lora}; \method{} (\MPname{}) is the full system of Section~\ref{sec:method}.

\paragraph{Metric and defaults.}
Recovery rate measures the fraction of the mentor--student gap closed: $\mathrm{Rec}=(\mathrm{score}-\mathrm{score}(\sonly))/(\mathrm{score}(\monly)-\mathrm{score}(\sonly))\times 100\%$. Values are not truncated; macro averages include datasets whose anchor gap exceeds one standard error (true for all runs). Unless stated otherwise, \MPname{} uses $R{=}16$ (Sections~\ref{sec:rsweep} and~\ref{sec:cost}); C2C, LoRA, and \method{} share the same training data and budget; all methods use greedy decoding with reasoning disabled, averaged over three seeds. Appendix~\ref{app:training} gives templates, trainable capacities, and search ranges.

% ---- Experiment 1: effectiveness histograms (double column) ----
\begin{figure*}[t]
\centering
\includegraphics[width=\textwidth]{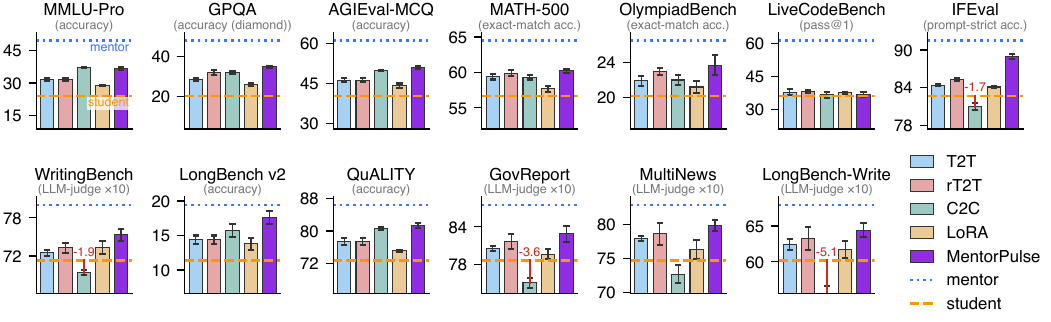}
\caption{Results on the thirteen datasets for A2; \MPname{} uses $R{=}16$, and rT2T refreshes the T2T guidance text on the same schedule (Appendix~\ref{app:rt2t}). Dotted/dashed lines are the \monly{}/\sonly{} anchors; error bars are standard deviations over three seeds; each y-axis is clipped to its anchor range.}
\label{fig:main}
\end{figure*}

\subsection{Overall Evaluation}
\label{sec:mainresults}

Figure~\ref{fig:main} gives the per-dataset picture on A2, where the anchors sit 6.3 to 27.7 points apart. \MPname{} closes 52.2\% of that gap on macro average, ahead of T2T at 26.9\%, LoRA at 17.5\%, and C2C at 10.9\%. rT2T refreshes the guidance text every 16 tokens under a protocol matched to \MPname{} (Appendix~\ref{app:rt2t}); it lifts macro recovery only to 33.7\%, significant over T2T on just two triggering tasks (Appendix~\ref{app:rt2t-quality}), while paying 41--54$\times$ \MPname{}'s per-request cost on the long-output representatives (Appendix~\ref{app:bytes}): refreshing in text space points the right way but carries little, and the step to 52.2\% belongs to the latent interface. The ordering is not uniform: on the five selection datasets (A--C and I--J) the two latent methods are close (\MPname{} 58.6\%, C2C 50.1\%): outputs of a few tokens leave static guidance no time to go stale. On the seven registered long-generation tasks (Table~\ref{tab:app-datasets}) the picture splits. On the five open-ended sets (G, H, K--M), C2C lands 1.7 to 5.1 points below the student line, clearing twice the pooled standard deviation on four of the five (on MultiNews the deficit is direction-consistent but within noise; the red drops in Figure~\ref{fig:main} mark the significant flips). On the two math sets (D, E) C2C keeps small positive gains, in line with the staleness account of Section~\ref{sec:diagnosis}: static guidance turns harmful where the target state evolves with the generated text, whereas a math problem statement keeps determining the target throughout the derivation. Averaged over the seven registered tasks, C2C recovery is $-15.8\%$ against \MPname{}'s 54.7\% ($-34.5\%$ versus 56.5\% on the five flipped sets). LiveCodeBench is the exception: no method recovers more than 7\% of the gap, matching the capability limit in Section~\ref{sec:diagnosis}.

% ---- Experiment 2: quadrant panels ----
\begin{figure}[t]
\centering
\includegraphics[width=\columnwidth]{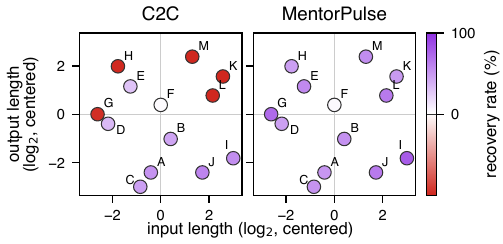}
\caption{The thirteen datasets in one length coordinate system; color is each panel's recovery rate, clipped to $[-20, 100]$, red marking harm. Coordinates are spread order-preservingly for readability (distances not to scale); point positions are identical in both panels.}
\label{fig:quadrant}
\end{figure}

Figure~\ref{fig:quadrant} re-plots the same scores in a shared length coordinate: $x$ and $y$ are each dataset's median input and output token counts ($\log_2$, centered on the median dataset), outputs measured from \monly{} generations. In the left panel C2C fades and turns red as output length grows: every red point (G, H, K, L, M) lies in the upper half. In the right panel the upper half stays purple and the lower half matches C2C closely; short outputs rarely reach a refresh boundary, so \MPname{} operates there as a static bridge. The \MPname{} upper half remains slightly lighter than its lower half: refresh reduces but does not remove the difficulty of long generation. The code point F stays near white in both panels.

\subsection{Ablations}
\label{sec:ablation}

\begin{table}[t]
\centering
\caption{Ablations on A2 at $R{=}16$, on two long-output sets (G, K) and two short-answer controls (A, J); $\Delta$Rec.\ is the change in mean recovery over the four sets relative to row 1. Rows 4 and 11 should stay close to row 1, whereas row 7 must reproduce \sonly{}; standard deviations in Appendix~\ref{app:full-results}.}
\label{tab:ablation}
\footnotesize
\setlength{\tabcolsep}{3.4pt}
\begin{tabular}{@{}cl cccc r@{}}
\toprule
\# & Ablation & G & K & A & J & $\Delta$Rec. \\
\midrule
1 & Full system ($R{=}16$) & 88.9 & 82.9 & 36.7 & 81.3 & --- \\
\addlinespace[2pt]
\multicolumn{7}{@{}l}{\textit{Refresh mechanism}} \\
2 & No refresh ($R\to\infty$) & 79.4 & 77.6 & 36.7 & 81.3 & $-41.9$ \\
3 & Refresh w/o WRT & 80.6 & 79.8 & 36.7 & 81.3 & $-32.3$ \\
4 & Full re-prefill & 88.6 & 83.0 & 36.7 & 81.3 & $-0.6$ \\
\addlinespace[2pt]
\multicolumn{7}{@{}l}{\textit{Memory content}} \\
5 & Mismatched mem. & 79.0 & 77.5 & 23.8 & 72.5 & $-72.1$ \\
6 & Random mem. & 81.5 & 78.6 & 23.8 & 72.3 & $-62.3$ \\
7 & Gates zeroed & 82.7 & 78.6 & 24.1 & 72.8 & $-57.7$ \\
\addlinespace[2pt]
\multicolumn{7}{@{}l}{\textit{Memory construction}} \\
8 & No tail slots & 85.1 & 81.6 & 36.0 & 80.3 & $-17.0$ \\
9 & Mean-pooled slots & 87.6 & 82.0 & 36.4 & 80.6 & $-7.8$ \\
10 & Uniform read-out & 88.2 & 82.0 & 36.2 & 80.7 & $-6.1$ \\
11 & Top-16 layers & 88.5 & 82.6 & 36.5 & 80.9 & $-2.9$ \\
\addlinespace[2pt]
\multicolumn{7}{@{}l}{\textit{Refresh details}} \\
12 & Two seg.\ types & 87.9 & 81.4 & 36.7 & 81.1 & $-7.4$ \\
\bottomrule
\end{tabular}
\end{table}

Table~\ref{tab:ablation} removes one design at a time. Removing refresh (row 2) is the largest mechanism-side drop: mean recovery on the two long-output sets falls from 59.4\% to $-24.4\%$, while the short-answer columns do not move a cell: those outputs end before the first refresh boundary. Row 3 separates schedule from training: swapping fresh memory into a bridge without windowed refresh training recovers little ($+1.2$ and $+2.2$ over the static row), and the training contributes the rest ($+8.3$ and $+3.1$); on IFEval both steps clear twice the pooled standard deviation, whereas on the LLM-judged GovReport they are direction-consistent but within noise, so the mechanism attribution reads from IFEval: the bridge must learn to digest mid-generation memory changes. Row 4 matches the full system: incremental refresh loses nothing against full recomputation, and Section~\ref{sec:cost} prices its cost advantage. The content controls repeat the attribution logic of Section~\ref{sec:diagnosis}: mismatched memory falls below the floor, random memory hovers near it, and zeroed gates reproduce \sonly{} bit for bit. The construction ablations (rows 8--10, 12) each cost several recovery points, tail slots mattering most on IFEval. Row 11 is designed not to drop: the top-16-layer read-out stays within noise on all four sets, making the pruned-layer transmission of Section~\ref{sec:cost} viable.

\subsection{Generality and Applicability}
\label{sec:analysis}

\begin{table}[t]
\centering
\caption{Macro-average recovery (\%) over the thirteen datasets by mentor group; best per row in bold. $N_M/N_S$ is the parameter ratio; \MPname{} uses $R{=}16$; averages follow Section~\ref{sec:setup-exp}; standard deviations in Appendix~\ref{app:full-results}; model names in Figure~\ref{fig:rd}.}
\label{tab:generality}
\footnotesize
\setlength{\tabcolsep}{4.5pt}
\begin{tabular}{@{}lr rrrr@{}}
\toprule
Pair & $N_M/N_S$ & T2T & C2C & LoRA & \MPname{} \\
\midrule
A1 (27B$\to$9B) & 3.0 & 25.3 & 9.2 & 16.3 & \textbf{54.0} \\
A2 (27B$\to$4B) & 6.8 & 26.9 & 10.9 & 17.5 & \textbf{52.2} \\
A3 (27B$\to$2B) & 13.5 & 18.9 & 8.7 & 11.8 & \textbf{35.7} \\
\midrule
B1 (31B$\to$12B) & 2.6 & 18.8 & 7.1 & 16.9 & \textbf{42.1} \\
\midrule
C1 (32B$\to$8B) & 4.0 & 25.5 & 11.9 & 14.9 & \textbf{54.6} \\
C2 (32B$\to$4B) & 8.2 & 22.4 & 11.4 & 16.0 & \textbf{45.8} \\
C3 (32B$\to$1.7B) & 19.3 & 14.5 & 5.9 & 9.6 & \textbf{26.8} \\
\midrule
D1 (14B$\to$8B) & 1.8 & 28.6 & 12.3 & 15.6 & \textbf{57.1} \\
D2 (14B$\to$4B) & 3.7 & 25.2 & 13.7 & 16.0 & \textbf{42.7} \\
D3 (14B$\to$1.7B) & 8.7 & 14.8 & 8.5 & 11.0 & \textbf{27.8} \\
\midrule
E1 (14B$\to$8B) & 1.8 & 26.4 & 11.3 & 13.9 & \textbf{53.0} \\
\bottomrule
\end{tabular}
\end{table}

% ---- Experiment 5: V64 diagnostic scatter ----
\begin{figure}[t]
\centering
\includegraphics[width=\columnwidth]{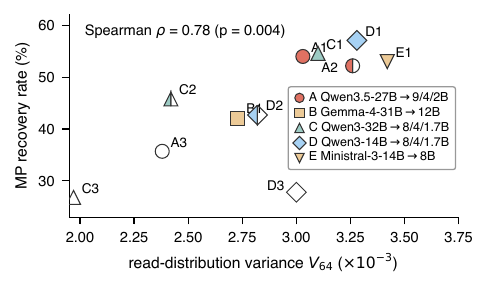}
\caption{Read-distribution variance $V_{64}$ against \MPname{} recovery (values as in Table~\ref{tab:generality}); marker and color encode the mentor group. Only the rank correlation is reported.}
\label{fig:rd}
\end{figure}

Table~\ref{tab:generality} shows the result survives a change of models: \MPname{} has the highest macro-average recovery on all eleven pairs and in all three families. The margins are smallest on C3 and D3, the most capability-separated pairs of their mentor groups: within each group, recovery falls monotonically as the parameter ratio grows (e.g., D1 57.1\% to D3 27.8\%).

During decoding, each bridge layer forms an attention distribution over the memory slots. $V_{64}$ is the variance of this read distribution, averaged over heads, layers, the first 64 decoded steps, and 32 samples per dataset, computable in minutes of GPU time per pair. High variance means selective reading; low variance means attention spread almost uniformly, diluting the guidance. Recovery tracks $V_{64}$ (Spearman $\rho = 0.78$, $p = 0.004$; Figure~\ref{fig:rd}), and the weakest pair, C3, has the lowest value---consistent with the reading that when the gap is too large the student cannot tell which slots matter, and reading degrades toward an average over the memory. Three limits apply: eleven pairs are observational evidence, $V_{64}$ is comparable only under one training recipe, and the check screens pairs rather than replacing evaluation.

\subsection{Refresh Interval}
\label{sec:rsweep}

% ---- Experiment 6: R sweep ----
\begin{figure}[t]
\centering
\includegraphics[width=\columnwidth]{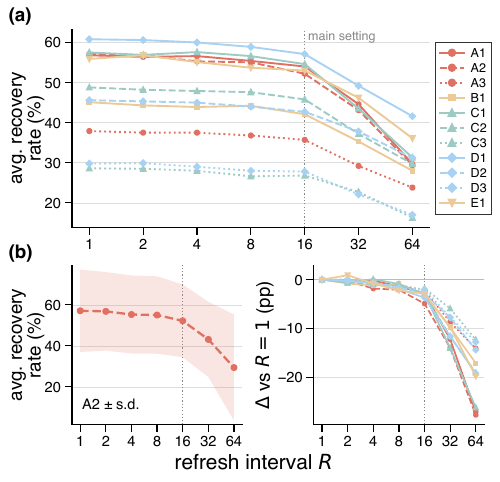}
\caption{Refresh-interval sweep at inference. (a) Macro-average recovery for all eleven pairs (exclusion rule of Section~\ref{sec:setup-exp}); the dotted line marks $R{=}16$. (b) A2 with its cross-dataset band, and change relative to $R{=}1$ in points.}
\label{fig:rsweep}
\end{figure}

Figure~\ref{fig:rsweep} sweeps $R\in\{1,2,4,8,16,32,64\}$ at inference. All eleven curves share one shape: nearly flat for $R\le 8$ (every pair within 2.2 recovery points of its $R{=}1$ value, small non-monotonic wobbles inside noise), declining from $R{=}16$, and down 12 to 28 recovery points by $R{=}64$; the shape is a property of the method: the plateau says guidance stays fresh for around ten tokens, so refreshing faster buys nothing. The main setting $R{=}16$ sits just past the plateau by intent: it gives up at most 4.9 recovery points against $R{=}1$, with long-output scores within one pooled standard deviation of $R{=}8$, and Section~\ref{sec:cost} places it among the cost-efficient settings $\{8,16,32\}$. $R{=}8$ stays among them but doubles the synchronizations, raising long-output per-request cost 35--50\%; at least one long-output task loses cost dominance at $R\leq4$. Quality-sensitive deployments with spare bandwidth can prefer $R{=}8$.

\subsection{Cost Accounting}
\label{sec:cost}

% ---- Experiment 7: cost dominance bars ----
\begin{figure}[t]
\centering
\includegraphics[width=\columnwidth]{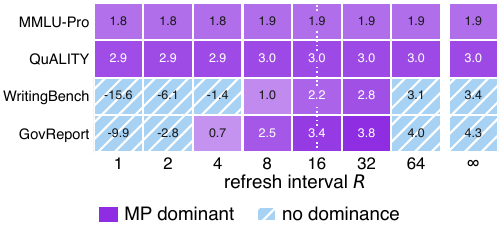}
\caption{Cost--quality dominance against T2T on A2. Purple: \MPname{} dominates; hatched: neither. Cell numbers give the per-request saving $\mathrm{Cost}_{\mathrm{T2T}}-\mathrm{Cost}_{\MPname}$ in $10^{-3}$ dollars; the dotted line marks $R{=}16$. rT2T is omitted, cost-dominated at every evaluated $R$ (Table~\ref{tab:app-rt2t-cost}).}
\label{fig:cost}
\end{figure}

We price with measured throughput, not theoretical FLOPs: prefill is compute-bound while decoding is bandwidth-bound, so FLOP counting understates decoding. Per-request cost sums each model's measured prefill and decode time at its GPU's rental rate, plus, for \MPname{}, the time to ship $\Bytes(R)$ (Eq.~\ref{eq:bytes}) per synchronization (formulas in Appendix~\ref{app:bytes}). Two structural facts follow: only the synchronization term grows as $R$ shrinks, since mentor compute is $R$-independent (Section~\ref{sec:refresh}); and T2T pays for mentor decoding of the guidance text, the most expensive per-token operation, which \MPname{} never performs---rT2T pays it at every refresh and never reaches a dominance panel (Table~\ref{tab:app-rt2t-cost}). One $R{=}16$ refresh with 16 transmitted layers ships 2.7\,MB in a measured 165\,ms, versus ${\sim}268$\,MB for a fused KV cache at a 2K-token input.

For each $R$ and the static limit $\infty$, we compare (quality, cost) by dominance---at least as good on both, strictly better on one---with no weighted score. Figure~\ref{fig:cost} covers four representative tasks, one per length type, under blocking cross-site synchronization (billing and sensitivities in Appendix~\ref{app:bytes}). On the long-output tasks \MPname{} dominates T2T at the evaluated settings $R\in\{8,16,32\}$ (GovReport already at $R{=}4$), $R{=}16$ in the middle; smaller $R$ keeps the highest quality but can lose cost dominance to synchronization overhead, and $R{=}64$ and $\infty$ are cheaper but no longer better. The short-output rows stay purple at every segment including $\infty$, with savings barely changing with $R$: refresh never triggers there, so the advantage is the static bridge's and we do not credit it to refresh. T2T-dominant segments never occur: text guidance always pays the mentor's decoding tax, so \MPname{} is strictly cheaper and can lose only on quality.

\section{Conclusion}
\label{sec:conclusion}

One-shot latent guidance does not stay reliable over long-form generation in our evaluated settings: static guidance falls below the student-only floor on long outputs, and a zero-training prefix-conditioned swap repairs it without touching the student's weights or cache. \method{} keeps guidance fresh through versioned slot memory, incremental mentor prefill, and windowed refresh training; across thirteen datasets and eleven pairs, refresh survives ablation as the main source of long-output gain and measured costs give the interval a defensible range. The gains require long outputs (short outputs never refresh; that advantage is the static bridge's), basic student capability, and a pair close enough for selective reading, screened by $V_{64}$. Guidance freshness is thus a design axis alongside capacity; learned triggers, prefill-free handoff, and cross-tokenizer pairs are future work.

\clearpage
\bibliographystyle{plainnat}
\bibliography{reference}

\clearpage
\appendix
\section{Experimental Setup}
\label{app:setup}

\subsection{Dataset Registry}
\label{app:datasets}

Table~\ref{tab:app-datasets} lists the thirteen evaluation datasets with the letter codes used throughout the paper, their metrics, and the evaluated subset sizes. LiveCodeBench (F) is restricted to problems released after July 2025, past the training cutoff of every evaluated model, as a contamination guard; QuALITY (J) uses the public dev split because test labels are not released. The long-generation suite (D, E, G, H, K, L, M) collects the generation-format tasks whose measured output medians span multiple refresh intervals (Appendix~\ref{app:lengths}); GPQA (B) is excluded as a selection task with only a brief justification (median 92), and LiveCodeBench (F), which qualifies by length (median 520), is excluded as the registered capability-limited exception (Section~\ref{sec:mainresults}).

\begin{table*}[!tbp]
\centering
\caption{The thirteen evaluation datasets. Letter codes are fixed throughout the paper; the last column marks the long-generation suite (inclusion rule in the text; length profiles in Table~\ref{tab:app-lengths}).}
\label{tab:app-datasets}
\small
\setlength{\tabcolsep}{5pt}
\begin{tabular}{@{}clllc@{}}
\toprule
Code & Dataset & Metric & Eval.\ subset & Long-gen.\ suite \\
\midrule
A & MMLU-Pro & accuracy & 2{,}000 (stratified) & --- \\
B & GPQA & accuracy (diamond) & 198 & --- \\
C & AGIEval-MCQ & accuracy & 2{,}000 (stratified) & --- \\
D & MATH-500 & exact-match accuracy & 500 & \ding{51} \\
E & OlympiadBench & exact-match accuracy (OE-TO-maths-en) & 675 & \ding{51} \\
F & LiveCodeBench & pass@1 & 400 (post 2025-07) & --- \\
G & IFEval & prompt-level strict accuracy & 541 & \ding{51} \\
H & WritingBench & LLM-judge mean $\times$10 & 1{,}000 & \ding{51} \\
I & LongBench v2 & accuracy & 503 & --- \\
J & QuALITY & accuracy & 2{,}086 (dev) & --- \\
K & GovReport & LLM-judge mean $\times$10 & 200 & \ding{51} \\
L & MultiNews & LLM-judge mean $\times$10 & 200 & \ding{51} \\
M & LongBench-Write & LLM-judge mean $\times$10 & 120 & \ding{51} \\
\bottomrule
\end{tabular}

\end{table*}

\subsection{Length Profiles and the Short-Answer Protocol}
\label{app:lengths}

Table~\ref{tab:app-lengths} registers the length profile behind the task-length map of Figure~\ref{fig:quadrant}. Input length is the median token count over evaluation inputs; output length is the median over \monly{} generations, since the mentor defines the target behavior; $L_g$ is the median T2T guidance length on the four datasets where the template was tuned. All medians are computed with the Qwen3.5 tokenizer; re-measuring with the Gemma-4 and Ministral~3 tokenizers changes medians by 3.1\% and 3.8\% respectively and moves no dataset across a length bin.

\begin{table}[!htbp]
\centering
\caption{Length profiles (median token counts). Outputs are measured from \monly{} generations; the tokenizer is Qwen3.5's (cross-family deviation ${<}4\%$).}
\label{tab:app-lengths}
\small
\setlength{\tabcolsep}{4pt}
\begin{tabular}{@{}clrrr@{}}
\toprule
Code & Dataset & Input med.\ & Output med.\ & $L_g$ med.\ \\
\midrule
A & MMLU-Pro & 410 & 6 & 130 \\
B & GPQA & 540 & 92 & --- \\
C & AGIEval-MCQ & 320 & 5 & --- \\
D & MATH-500 & 140 & 430 & --- \\
E & OlympiadBench & 280 & 570 & --- \\
F & LiveCodeBench & 470 & 520 & --- \\
G & IFEval & 110 & 486 & --- \\
H & WritingBench & 160 & 940 & 250 \\
I & LongBench v2 & 9,400 & 7 & --- \\
J & QuALITY & 5,100 & 6 & 205 \\
K & GovReport & 7,600 & 705 & 310 \\
L & MultiNews & 6,300 & 566 & --- \\
M & LongBench-Write & 4,300 & 1,080 & --- \\
\bottomrule
\end{tabular}

\end{table}

\paragraph{Short-answer protocol.}
The four pure multiple-choice sets (A, C, I, J) are evaluated under a short-answer protocol: in no-think mode the student is asked to output only the option letter, the standard treatment of choice tasks in lm-eval-style harnesses. Their output medians are 5--7 tokens, so at the main setting $R{=}16$ these datasets literally never reach a refresh boundary: \mpat{16} operates as the static bridge on them. This single registered fact grounds three places in the paper: the ``$=$static'' cells of the $R$ sweep (Table~\ref{tab:app-rsweep}), the unchanging short-output rows of Figure~\ref{fig:cost}, and the applicability statement of Section~\ref{sec:analysis}. GPQA (B) keeps a brief free-form justification (median 92 tokens) and serves as the transitional sample: refresh triggers a few times with near-zero benefit. MATH-500 (D) and OlympiadBench (E) span many refresh windows and are discussed with the long-output tasks.

\subsection{Runtime Environment and Availability}
\label{app:env}

Bridge checkpoints are named \texttt{fb6-<pair>-r16-e2} (eleven, one per pair), C2C fusers \texttt{c2c-fuser-<pair>-v1}, and LoRA adapters \texttt{lora-r64-<pair>}. Scoring uses lm-eval-harness 0.4.9 for the standard sets and our long-form evaluator longeval v2.1 for the judged sets (H, K, L, M and the judged diagnostic tasks), with GPT-4.1 as the fixed judge at temperature 0. All decoding is greedy with reasoning disabled (no-think), with per-dataset generation caps; seeds are $\{13, 42, 2026\}$. The three trainable objects share the training set fuse\_v3 (51{,}908 rows; Appendix~\ref{app:training}). Results were collected between June~2 and July~26, 2026. Every reported score, including the \monly{} and \sonly{} anchors, follows the protocol of Section~\ref{sec:setup-exp}: at least three seeds (Appendix~\ref{app:stats}), the full registered evaluation subsets of Table~\ref{tab:app-datasets}, and the deployment serving stack (vLLM 0.19.1 in bf16). One backend exception is registered: the Gemma-4 family (pair B1) is not supported by this vLLM build and is served through HuggingFace transformers in bf16 with identical decoding settings. Within every pair, anchors and all compared methods run on the same backend, so no comparison in the paper mixes serving stacks; the cost model prices only the A2 deployment pair and is unaffected. Separately, a lightweight screening pass over all thirteen models and datasets (July~26, 2026; \texttt{limit}=500, single run, same serving stack, Gemma-4 through HF) sanity-checked the capability ordering assumed in pair selection; no number from that pass enters any table, and recovery rates in particular divide method scores and anchor scores drawn from the same full-subset runs. Serving throughputs for the cost model are measured separately on the deployment stack (Appendix~\ref{app:bytes}). Code, evaluation configurations, the data pipeline, per-seed outputs, and the scripts that derive every table in this appendix will be released.

\subsection{Statistical Discipline}
\label{app:stats}

Every reported number is the mean over at least three seeds, written as mean$\pm$standard deviation; methods that train vary the training seed, inference-only conditions vary the evaluation seed. When two conditions $a, b$ are compared, the pooled standard deviation is $\sigma_{ab} = \smash{\sqrt{\sigma_a^2 + \sigma_b^2}}$, and any difference below $2\sigma_{ab}$ is described as ``comparable'' rather than better or worse; every claim of significance in the paper passes this threshold. Recovery is $(\mathrm{score} - \sonly{})/(\monly{} - \sonly{}) \times 100\%$, reported without truncation (negative and ${>}100\%$ values kept; only figure color scales clip, as stated in their captions). Datasets whose mentor--student anchor gap is below one standard error would be excluded from macro-average recovery as unstable; no evaluated pair triggered this rule.

% ======================================================================
\section{Condition Definitions and Paradigm Properties}
\label{app:conditions-sec}

\subsection{Experimental Conditions}
\label{app:conditions}

Table~\ref{tab:app-conditions} defines every condition used in the paper. Main-text performance results use \sonly{}, \monly{}, T2T, C2C, LoRA, and \mpat{16}; the diagnosis section (Section~\ref{sec:diagnosis}) uses \oneshot{}, \oracleswap{16}, and the two mismatch controls; rT2T appears beside T2T in Figure~\ref{fig:main} and in Appendices~\ref{app:rt2t-quality} and~\ref{app:bytes}; the remaining rows are ablation controls (Table~\ref{tab:ablation}).

\begin{table*}[!tbp]
\centering
\caption{All experimental conditions. ``Mentor decodes'' refers to inference time only; offline label generation is excluded. \oneshot{} and the ablation row MP$-$refresh are the same condition under two names (static bridge, $R{\to}\infty$).}
\label{tab:app-conditions}
\footnotesize
\setlength{\tabcolsep}{4pt}
\begin{tabular}{@{}l p{8.4cm} p{3.2cm} l@{}}
\toprule
Condition & Definition & Role & Mentor decodes \\
\midrule
\sonly{} & student alone, no bridge, no memory & floor & --- \\
\monly{} & mentor answers alone & ceiling & yes \\
T2T & one-shot mentor guidance text ($L_g \le 384$) prepended to the student prompt & text-interface route & yes (once) \\
rT2T@$R$ & T2T guide refreshed every $R$ tokens; replace-style rebuild, delimiter-stripped & text $\times$ refresh cell (Appendix~\ref{app:rt2t}) & yes (every refresh) \\
C2C & mentor layer-wise KV fused into the student cache once before decoding & static-latent route & no \\
LoRA & equal-budget low-rank finetuning, no memory & parameter route & --- \\
\mpat{R} & full method after WRT, refresh interval $R$ & main method ($R{=}16$) & no \\
\MPname{}-ST & \MPname{} with a learned staleness trigger & stage-2 outlook (Appendix~\ref{app:trigger}) & no \\
\oneshot{} & bridge with static memory ($R{\to}\infty$; $=$ MP$-$refresh) & one-shot guidance instance & no \\
\oracleswap{R} & zero-training rebuild of the memory from $x \oplus \hat{y}_{<t}$, swapped every $R$ tokens & staleness diagnosis & no \\
\oracleswapmis{R} & initial memory matched; refresh points swap in \emph{fresh} memory of another sample & refresh $\times$ content control & no \\
\oneshot{}-mismatch & initial memory taken from another sample, never refreshed & static content canary & no \\
Random & random-vector memory of the correct shape & information control & no \\
Zero / gate-off & zero memory or closed gates & degeneracy check ($=$ \sonly{}, bit-exact) & no \\
\bottomrule
\end{tabular}
\end{table*}

\subsection{Paradigm Properties}
\label{app:paradigms}

Table~\ref{tab:paradigms} compares the closest cross-model interfaces and lossless speculative decoding on the three properties discussed in Section~\ref{sec:related}: whether the mentor stays frozen, whether the transferred state count is explicitly capped, and whether the transferred signal is refreshed from the generated prefix. Speculative decoding is included as a discussion-only contrast: it is lossless and couples the two models token by token on one machine, so quality and cost are not comparable with the lossy, sparsely-synchronized interfaces above it.

\begin{table}[!htbp]
\centering
\caption{Property comparison with the closest cross-model interfaces and speculative decoding. Refresh means updating transferred guidance from the generated prefix; dashes mark properties that do not apply.}
\label{tab:paradigms}
\small
\setlength{\tabcolsep}{2.5pt}
\begin{tabular}{l@{\hspace{4pt}}ccccc}
\toprule
Property & L2S & C2C & LGR & \specdec{} & \textbf{\MPname{}} \\
\midrule
Mentor frozen & \ding{51} & \ding{51} & \ding{55} & \ding{51} & \ding{51} \\
Explicit state-count cap & \ding{55} & \ding{55} & \ding{51} & --- & \ding{51} \\
Prefix-conditioned refresh & \ding{55} & \ding{55} & \ding{55} & --- & \ding{51} \\
Matches mentor distribution & \ding{55} & \ding{55} & \ding{55} & \ding{51} & \ding{55} \\
\bottomrule
\end{tabular}
\end{table}

% ======================================================================
\section{Baseline Implementations and Fairness}
\label{app:baselines}

Each baseline below states its implementation and the hyperparameter search it received, so that no comparison rests on an under-tuned opponent.

\subsection{T2T}
\label{app:t2t}

The guidance template \texttt{t2t-guide-v3} instructs the mentor to read the input and produce, in order, a task analysis, the key points a correct answer must satisfy, and a short answer plan; the guidance is truncated at $L_g \le 384$ tokens and prepended to the student prompt. The template went through three iterations on held-out development data (v1: free-form summary; v2: added the key-point list; v3: added the answer plan and tightened the length cap), and v3 was selected by development-set student score.

\subsection{rT2T}
\label{app:rt2t}

rT2T is the refreshed version of T2T: it fills the ``text $\times$ refresh'' cell of the $\{$text, latent$\} \times \{$static, refreshed$\}$ matrix, so that the refresh schedule and the representation format are separated as factors. Its protocol is fixed by five clauses, each of which protects a specific fairness property:
\begin{enumerate}
\item \textbf{Mentor conditioning.} At each refresh the mentor incrementally prefills $[x; \hat{y}_{<t}]$, the same fresh path \method{} uses, and decodes a new hint of $L_g'$ tokens (template \texttt{t2t-hint-v1}).
\item \textbf{Delimiters.} Hints are injected wrapped in \texttt{<guide>}\dots\texttt{</guide>} and stripped, delimiters and content, before scoring, so no hint text can leak into the graded output.
\item \textbf{Replace-style injection (main setting).} Each refresh discards the previous hint (including the initial guide) and rebuilds $[\mathrm{hint}_j; x; \hat{y}_{<t}]$ with a full re-prefill. Replacement is strictly more expensive than accumulation in every (task, $R$, $L_g'$) cell of Table~\ref{tab:app-rt2t-cost}---by a second-order margin, about 2\% of total cost---and keeps the context clean, mirroring \method{}'s replacement of stale memory; we adopt the more expensive, cleaner variant as the most generous implementation. The accumulation variant is reported as a robustness check in Appendix~\ref{app:rt2t-quality}.
\item \textbf{Hint budget.} $L_g' \in \{64, 128\}$ is swept and the per-task best is reported (best-effort baseline).
\item \textbf{Initial guide $=$ T2T.} This yields two identity checks for free: on short-answer tasks (output ${<}R$, $n_{\mathrm{sync}}{=}0$) and at $R{\to}\infty$, rT2T must equal T2T cell by cell. Both identities are hard-verified by the fill scripts (Appendix~\ref{app:consistency}).
\end{enumerate}
rT2T is the rigorous return of the retired TH-$k$ condition of an earlier design: TH-$k$ appended hints on a fixed schedule without conditioning the mentor on $\hat{y}_{<t}$, without replacement, and without delimiter stripping, and is superseded in all experiments.

\subsection{C2C}
\label{app:c2c}

The C2C reproduction follows the layer-wise cache-fusion design of \citet{fu2026c2c}: the mentor's per-layer KV cache for the input is projected and fused into the student's cache by a trainable fuser before decoding, and never updated afterwards. The fuser (\texttt{c2c-fuser-<pair>-v1}) is trained on the same data and budget as the bridge; its width is chosen per pair so that its trainable parameter count matches the bridge's, and its learning rate is searched over the same grid as \method{}'s.

\subsection{LoRA}
\label{app:lora}

The LoRA control uses rank 64 on all pairs, with adapters on the attention and MLP projections. Trainable parameter counts are matched to the bridge pair by pair, ranging from 21M to 58M across the eleven pairs, so any gain of \method{} over LoRA cannot be attributed to extra capacity. The learning rate is searched over the shared grid; training data and budget are identical to the bridge's.

\subsection{Fairness Summary}
\label{app:fairness}

The three trainable objects (C2C fuser, LoRA adapters, \method{} bridge) use the same training set fuse\_v3, the same step budget, and the same learning-rate search; all methods are served with the same stack, decode greedily without thinking mode, and are scored by the same per-dataset script; every number averages at least three seeds (Appendix~\ref{app:stats}). rT2T contains no trainable object; its fairness rests on the five-clause protocol above.

% ======================================================================
\section{Training Details}
\label{app:training}

\subsection{Training Data}
\label{app:data}

Two offline passes prepare the training store. The first pass generates long targets: prompts are drawn from open instruction sources for writing, explanation, and rewriting, code instructions in the OSS-Instruct style \citep{wei2024magicoder}, synthetic instruction-following prompts with verifiable constraints \citep{zhou2023ifeval}, and long document-grounded targets. The mentor decodes greedily without thinking mode, up to 768 new tokens; this is the only mentor decoding in the entire pipeline and happens once, offline. Four quality gates screen the generations: a repetition gate (4-gram), a completeness gate (natural EOS within a length window), a language gate, and a code gate rejecting broken fenced blocks. Each gate removes 5--15\% of the rows it screens; the failures overlap, and together the four gates drop 24.4\% of prompts. For the 27B mentor, 26{,}997 prompts yield 20{,}408 long labels (75.6\% kept). The second pass extracts mentor states: prefill only, all layers, written under the slot layout to a store of about 1.9\,TB in bf16. The mixed set fuse\_v3 combines all 20{,}408 filtered long targets (39.3\%) with 31{,}500 short-answer rows downsampled from the static stage (60.7\%), 51{,}908 rows in total; multiple-choice rows restrict the answer alphabet to the true option count of each row to remove a leakage artifact. All sources are disjoint from the thirteen evaluation sets and from the diagnostic suite (Appendix~\ref{app:dsuite}), so no refresh gain can come from memorized evaluation data.

\subsection{Windowed Refresh Training}
\label{app:wrt}

WRT samples a refresh point $t$ per row (at the $R{=}16$ grid, including $t{=}0$, which keeps static training inside the objective) and computes the loss only on the window $(t, t{+}R]$; the prefix enters as the source of the refreshed memory, never as supervision. By causality, one extraction pass over prompt-plus-label provides the mentor states for every candidate refresh time, so storage grows only linearly (about 768 extra tokens per row). An optional second round replaces a subset of label prefixes with prefixes sampled from the bridged student, following dataset aggregation \citep{ross2011dagger}, to shrink the train--test prefix mismatch stated in Section~\ref{sec:method}.

\subsection{Optimization and Budget}
\label{app:optim}

Both backbones, embeddings, and heads are frozen; the trained parameters are the shared projector, the per-layer rank-64 corrections, the layer, position, and segment embeddings, the read gates $\rho_o$, and the per-layer cross-attention blocks (rank 256). Training runs 2 epochs at learning rate $2\times10^{-5}$ (selected on the development split from the grid shared with C2C and LoRA) with answer-segment cross-entropy, weighted by token share. The bridge warm-starts from the static-stage checkpoint with gates not reset; inherited positions map to the tail segment of the extended position table and all new entries initialize at zero, so the extended system is bit-identical to the inherited one at step zero. Microbatches are grouped by a slot budget of 256 rather than a row count, because the injected memory is projected to K/V at every injection layer and retained for backward; gradient checkpointing is disabled as incompatible with the forward hooks that mount the bridge. Generation health during development is monitored with distinct-2, 4-gram repetition, natural-EOS rate, and mean length; alarms fire below 90\% EOS rate or above 10\% repetition. Table~\ref{tab:app-training} consolidates the pipeline statistics.

\begin{table}[!htbp]
\centering
\caption{Training pipeline and hyperparameter summary.}
\label{tab:app-training}
\small
\setlength{\tabcolsep}{3pt}
\begin{tabular}{ll}
\toprule
Item & Value \\
\midrule
Long-target prompts / kept labels & 26{,}997 / 20{,}408 \\
Quality-gate filter rate & 5--15\% per gate, 24.4\% total \\
Mixed training rows (fuse\_v3) & 51{,}908 \\
Long / short-answer mix & 20{,}408 / 31{,}500 (39\% / 61\%) \\
Mentor state store (27B, bf16) & ${\sim}1.9$\,TB \\
Max label length & 768 tokens \\
Learning rate / epochs & $2\times10^{-5}$ / 2 \\
Bridge rank / per-layer correction rank & 256 / 64 \\
Slot budget per microbatch & 256 \\
WRT refresh-point grid & $R{=}16$, incl.\ $t{=}0$ \\
EOS-rate alarm / repetition alarm & $<$90\% / $>$10\% \\
\bottomrule
\end{tabular}
\end{table}

% ======================================================================
\section{Method Implementation Details}
\label{app:method}

\subsection{Slot Memory Layout}
\label{app:slots}

For each selected mentor layer $o$, retained states are normalized by a fixed scalar $\mathrm{rms}_o$ and mapped from $d_m$ to $d_t$ by a shared projection with a rank-64 layer-specific correction. Segment slots summarize the prompt and the generated prefix, each taking the true hidden state at its segment's final position; the current tail keeps 32 raw-state slots for exact recent constraints. Prompt segments default to $S{=}32$ tokens under the cap $p_x \le P{=}128$: longer inputs are re-segmented into $P$ evenly coarser segments (roughly 73-token segments at the 9,400-token LongBench~v2 median input), which caps the initial transfer regardless of input length (Appendix~\ref{app:bytes}). Generated-prefix slots are appended at one slot per 32 tokens under a separate budget of 128 slots (4{,}096 output tokens) with oldest-first eviction as a backstop; prompt-summary slots are never evicted. No per-dataset generation cap (Appendix~\ref{app:env}) reaches this budget --- the longest registered output median, 1{,}080 tokens on LongBench-Write, fills 34 slots (Table~\ref{tab:app-lengths}) --- so the backstop never fires in our runs. Three segment-type embeddings distinguish prompt summary, generated-prefix summary, and current tail; layer and slot-order embeddings complete the layout, and a validity mask handles variable lengths. A learned weight $\rho_o$ softly combines mentor layers and defines the pruning order used in Section~\ref{sec:cost}. All evaluated pairs share a tokenizer family, so the mentor consumes student-generated tokens directly.

\subsection{Gated Cross-Attention Bridge}
\label{app:gxb}

Each student layer carries a gated cross-attention block whose query, key, value, and output projections have rank 256; the query is the student residual stream and the memory supplies keys and values, projected locally on the student side (the wire payload is one slot tensor, not separate K/V tensors). Scalar gates start at zero, so the untrained bridge is an identity map and the system reproduces the plain student bit for bit --- the degeneracy check used in Sections~\ref{sec:diagnosis} and~\ref{sec:ablation} and the gate-off row of Table~\ref{tab:app-conditions}. Training escapes this closed-gate dead zone by warm-starting from the static-stage checkpoint whose gates are already open, without resetting them; new position and segment entries initialize at zero so the inherited path is unchanged at the start of WRT.

\subsection{Versioned Incremental Refresh}
\label{app:refresh-impl}

At a refresh the mentor incrementally prefills only the $R$ newly generated tokens, appends the resulting segment slots, and rebuilds the tail; the memory tensor is replaced in place (hot swap) as a new version $\mem_{t_j}$. Because the memory is a side tensor rather than part of the student's self-attention cache, the student's KV cache and generated text are never invalidated; older summary slots are retained on the student side and are not retransmitted, so each refresh ships only the new summary slots and the current tail. Row 4 of Table~\ref{tab:ablation} verifies that this incremental path matches full recomputation within noise.

% ======================================================================
\section{Complete Experimental Results}
\label{app:full-results}

\subsection{Main Pair}
\label{app:main-pair}

Table~\ref{tab:app-main} gives the full per-dataset scores with standard deviations for the main pair A2, the source of Figure~\ref{fig:main}; Table~\ref{tab:app-ablation-sd} gives the raw scores behind the recovery-based ablation Table~\ref{tab:ablation}. On the sign-flip claim, C2C sits below \sonly{} on five of the seven registered long-generation tasks (G $-1.7$, H $-1.9$, K $-3.6$, M $-5.1$, each clearing twice the pooled standard deviation, thresholds 1.13--4.25; on MultiNews the deficit of $-2.0$ against a 2.86 threshold is direction-consistent but within noise, and the main text treats that set accordingly), and keeps small positive gains on the two math sets (D $+2.6$, E $+1.8$). No short-output or capability-limited set flips. In Table~\ref{tab:app-ablation-sd}, the A and J columns of rows 2--4 equal row 1 cell by cell: those outputs end before the first refresh boundary, so every initially-matched refresh condition is structurally identical to the full system there (the zero-trigger identity of Appendix~\ref{app:consistency}); row 5 drops on A and J because its \emph{initial} memory is already mismatched, a static-mismatch condition.

\begin{table*}[!tbp]
\centering
\caption{Full results for the main pair A2 (Qwen3.5-27B $\to$ Qwen3.5-4B), mean$\pm$sd over three seeds; \mpat{16} throughout. The macro-recovery row uses the definition of Appendix~\ref{app:stats}.}
\label{tab:app-main}
\small
\setlength{\tabcolsep}{4.5pt}
\begin{tabular}{@{}cl cccccc@{}}
\toprule
Code & Dataset & \monly{} & \sonly{} & T2T & C2C & LoRA & \mpat{16} \\
\midrule
A & MMLU-Pro & 49.5$\pm$0.1 & 24.1$\pm$0.2 & 31.6$\pm$0.7 & 37.0$\pm$0.1 & 28.9$\pm$0.2 & 36.7$\pm$0.8 \\
B & GPQA & 47.9$\pm$0.2 & 20.2$\pm$0.2 & 28.3$\pm$0.8 & 32.0$\pm$0.8 & 25.8$\pm$0.7 & 34.7$\pm$0.4 \\
C & AGIEval-MCQ & 61.1$\pm$0.1 & 40.4$\pm$0.0 & 46.2$\pm$0.7 & 49.8$\pm$0.2 & 44.1$\pm$0.9 & 51.0$\pm$0.4 \\
D & MATH-500 & 64.5$\pm$0.1 & 56.7$\pm$0.1 & 59.4$\pm$0.4 & 59.3$\pm$0.3 & 57.7$\pm$0.4 & 60.2$\pm$0.3 \\
E & OlympiadBench & 26.5$\pm$0.1 & 20.2$\pm$0.1 & 21.9$\pm$0.6 & 22.0$\pm$0.6 & 21.2$\pm$0.7 & 23.7$\pm$1.1 \\
F & LiveCodeBench & 61.3$\pm$0.2 & 36.3$\pm$0.2 & 38.0$\pm$1.2 & 36.8$\pm$1.4 & 37.6$\pm$0.5 & 37.0$\pm$1.2 \\
G & IFEval & 91.5$\pm$0.0 & 82.7$\pm$0.0 & 84.4$\pm$0.2 & 81.0$\pm$0.6 & 84.1$\pm$0.1 & 88.9$\pm$0.4 \\
H & WritingBench & 80.0$\pm$0.5 & 71.3$\pm$0.4 & 72.5$\pm$0.5 & 69.4$\pm$0.4 & 73.3$\pm$1.0 & 75.3$\pm$1.0 \\
I & LongBench v2 & 19.4$\pm$0.0 & 11.4$\pm$0.1 & 14.4$\pm$0.6 & 15.7$\pm$0.9 & 13.8$\pm$0.9 & 17.6$\pm$0.9 \\
J & QuALITY & 86.4$\pm$0.1 & 72.8$\pm$0.1 & 77.5$\pm$0.9 & 80.7$\pm$0.3 & 75.2$\pm$0.2 & 81.3$\pm$0.6 \\
K & GovReport & 87.5$\pm$0.5 & 78.6$\pm$1.4 & 80.5$\pm$0.4 & 75.0$\pm$0.8 & 79.6$\pm$0.8 & 82.9$\pm$1.3 \\
L & MultiNews & 82.8$\pm$0.4 & 74.7$\pm$0.6 & 77.9$\pm$0.4 & 72.7$\pm$1.3 & 76.3$\pm$1.4 & 79.8$\pm$0.8 \\
M & LongBench-Write & 67.9$\pm$0.6 & 60.2$\pm$1.4 & 62.4$\pm$0.8 & 55.1$\pm$1.6 & 61.7$\pm$1.2 & 64.4$\pm$1.0 \\
\midrule
\multicolumn{2}{@{}l}{Macro recovery (\%)} & --- & --- & 26.9 & 10.9 & 17.5 & 52.2 \\
\bottomrule
\end{tabular}

\end{table*}

\begin{table*}[!tbp]
\centering
\caption{Raw scores with standard deviations for the twelve ablation rows of Table~\ref{tab:ablation} (A2, $R{=}16$; two long-output sets G, K and two short-answer controls A, J). ``Reused'' rows are copied from their registered source, not re-evaluated.}
\label{tab:app-ablation-sd}
\footnotesize
\setlength{\tabcolsep}{4pt}
\begin{tabular}{@{}cl cccc l@{}}
\toprule
\# & Condition & G & K & A & J & Runs \\
\midrule
1 & MP (full system, $R{=}16$) & 88.9$\pm$0.4 & 82.9$\pm$1.3 & 36.7$\pm$0.8 & 81.3$\pm$0.6 & reused ($=$ \mpat{16}, Tab.~\ref{tab:app-main}) \\
\addlinespace[2pt]
\multicolumn{7}{@{}l}{\textit{Refresh mechanism}} \\
2 & MP$-$refresh ($R\to\infty$) & 79.4$\pm$0.2 & 77.6$\pm$0.9 & 36.7$\pm$0.8 & 81.3$\pm$0.6 & retrained \\
3 & MP$-$WRT (zero-training swap) & 80.6$\pm$0.5 & 79.8$\pm$0.9 & 36.7$\pm$0.8 & 81.3$\pm$0.6 & inference only \\
4 & MP$-$full (full re-prefill) & 88.6$\pm$0.2 & 83.0$\pm$0.9 & 36.7$\pm$0.8 & 81.3$\pm$0.6 & inference only \\
\addlinespace[2pt]
\multicolumn{7}{@{}l}{\textit{Memory content}} \\
5 & MP$-$mismatch (incl.\ initial) & 79.0$\pm$0.4 & 77.5$\pm$0.7 & 23.8$\pm$0.5 & 72.5$\pm$0.8 & inference only \\
6 & MP$-$random & 81.5$\pm$0.8 & 78.6$\pm$1.1 & 23.8$\pm$0.7 & 72.3$\pm$0.5 & inference only \\
7 & MP$-$gate0 & 82.7$\pm$0.0 & 78.6$\pm$1.4 & 24.1$\pm$0.2 & 72.8$\pm$0.1 & reused ($=$ \sonly{}); bit-identity passed \\
\addlinespace[2pt]
\multicolumn{7}{@{}l}{\textit{Memory construction}} \\
8 & MP$-$tail & 85.1$\pm$0.2 & 81.6$\pm$1.2 & 36.0$\pm$0.2 & 80.3$\pm$0.3 & retrained \\
9 & MP$-$meanpool & 87.6$\pm$0.3 & 82.0$\pm$1.0 & 36.4$\pm$0.5 & 80.6$\pm$0.8 & retrained \\
10 & MP$-$uniform & 88.2$\pm$0.3 & 82.0$\pm$0.8 & 36.2$\pm$0.4 & 80.7$\pm$0.2 & retrained \\
11 & MP$-$prune16 (top-16 layers) & 88.5$\pm$0.7 & 82.6$\pm$0.6 & 36.5$\pm$0.9 & 80.9$\pm$0.6 & inference only \\
\addlinespace[2pt]
\multicolumn{7}{@{}l}{\textit{Refresh engineering}} \\
12 & MP$-$2seg & 87.9$\pm$0.2 & 81.4$\pm$0.6 & 36.7$\pm$0.5 & 81.1$\pm$0.6 & retrained \\
\bottomrule
\end{tabular}

\end{table*}

\subsection{All Eleven Pairs}
\label{app:pairs}

Tables~\ref{tab:app-pairs-1}--\ref{tab:app-pairs-5} give the complete per-dataset results for the ten remaining pairs; A2 is in Table~\ref{tab:app-main}. The macro-recovery rows reproduce Table~\ref{tab:generality} and the $R{=}16$ column of Table~\ref{tab:app-rsweep-pairs}. Capability-limited registration: on B1, the Gemma-4-12B student scores near zero on MATH-500 and OlympiadBench, far below these tasks' difficulty floor. Consistent with the capability limit of Section~\ref{sec:diagnosis}, no comparison method recovers a meaningful fraction of such a gap, so the derived columns follow a capability-limited profile (near-zero gains for every method) and these two cells are excluded from the long-output gain narrative.

\begin{table*}[!tbp]
\centering
\caption{Full per-dataset results, pairs A1 and A3 (datasets by letter code, Table~\ref{tab:app-datasets}); mean$\pm$sd over three seeds.}
\label{tab:app-pairs-1}
\footnotesize
\setlength{\tabcolsep}{2.6pt}
\begin{tabular}{@{}c cccccc@{}}
\toprule
  & \monly{} & \sonly{} & T2T & C2C & LoRA & \mpat{16} \\
\midrule
\multicolumn{7}{@{}l}{\textbf{A1}: Qwen3.5-27B $\to$ Qwen3.5-9B ($N_M/N_S = 3.0$)} \\
\addlinespace[1pt]
A & 49.5$\pm$0.1 & 27.4$\pm$0.1 & 34.7$\pm$0.6 & 39.0$\pm$0.3 & 31.4$\pm$0.2 & 39.6$\pm$0.2 \\
B & 47.9$\pm$0.2 & 27.9$\pm$0.0 & 33.1$\pm$0.7 & 38.3$\pm$0.6 & 31.9$\pm$0.9 & 40.0$\pm$0.9 \\
C & 61.1$\pm$0.1 & 47.4$\pm$0.1 & 52.1$\pm$0.5 & 54.1$\pm$0.5 & 50.1$\pm$0.4 & 55.2$\pm$0.4 \\
D & 64.5$\pm$0.1 & 58.3$\pm$0.0 & 59.4$\pm$0.8 & 59.6$\pm$0.3 & 59.4$\pm$0.9 & 60.9$\pm$0.9 \\
E & 26.5$\pm$0.1 & 21.7$\pm$0.0 & 23.1$\pm$0.8 & 22.5$\pm$1.2 & 23.0$\pm$0.7 & 24.5$\pm$1.4 \\
F & 61.3$\pm$0.2 & 49.7$\pm$0.2 & 50.1$\pm$0.3 & 49.5$\pm$0.4 & 50.2$\pm$0.8 & 50.4$\pm$0.6 \\
G & 91.5$\pm$0.0 & 85.1$\pm$0.2 & 86.1$\pm$0.3 & 82.4$\pm$0.7 & 85.8$\pm$0.4 & 89.1$\pm$0.3 \\
H & 80.0$\pm$0.5 & 72.6$\pm$0.6 & 74.3$\pm$1.1 & 70.1$\pm$1.2 & 73.2$\pm$0.6 & 76.7$\pm$0.7 \\
I & 19.4$\pm$0.0 & 14.7$\pm$0.1 & 15.8$\pm$0.4 & 16.7$\pm$0.3 & 14.9$\pm$0.3 & 18.1$\pm$0.8 \\
J & 86.4$\pm$0.1 & 76.4$\pm$0.1 & 79.9$\pm$0.5 & 81.9$\pm$0.3 & 78.1$\pm$0.5 & 83.0$\pm$0.2 \\
K & 87.5$\pm$0.5 & 81.6$\pm$0.6 & 83.0$\pm$0.4 & 79.8$\pm$0.7 & 82.8$\pm$1.2 & 85.6$\pm$0.6 \\
L & 82.8$\pm$0.4 & 77.4$\pm$0.4 & 79.3$\pm$0.2 & 75.4$\pm$0.6 & 78.9$\pm$0.7 & 79.5$\pm$0.5 \\
M & 67.9$\pm$0.6 & 61.1$\pm$0.9 & 63.1$\pm$0.6 & 59.5$\pm$0.9 & 62.2$\pm$1.3 & 65.2$\pm$1.0 \\
\addlinespace[1pt]
\multicolumn{2}{@{}l}{Macro rec.\ (\%)} & --- & 25.3 & 9.2 & 16.3 & 54.0 \\
\bottomrule
\end{tabular}
\hfill
\begin{tabular}{@{}c cccccc@{}}
\toprule
  & \monly{} & \sonly{} & T2T & C2C & LoRA & \mpat{16} \\
\midrule
\multicolumn{7}{@{}l}{\textbf{A3}: Qwen3.5-27B $\to$ Qwen3.5-2B ($N_M/N_S = 13.5$)} \\
\addlinespace[1pt]
A & 49.5$\pm$0.1 & 16.0$\pm$0.0 & 23.3$\pm$0.3 & 27.3$\pm$0.5 & 20.5$\pm$0.4 & 28.3$\pm$0.2 \\
B & 47.9$\pm$0.2 & 19.9$\pm$0.1 & 25.6$\pm$1.3 & 29.6$\pm$1.2 & 23.0$\pm$0.8 & 30.8$\pm$0.3 \\
C & 61.1$\pm$0.1 & 32.6$\pm$0.0 & 38.0$\pm$0.9 & 42.0$\pm$0.1 & 35.3$\pm$0.5 & 42.5$\pm$0.8 \\
D & 64.5$\pm$0.1 & 51.4$\pm$0.0 & 53.8$\pm$0.5 & 54.3$\pm$0.4 & 52.9$\pm$0.6 & 56.1$\pm$0.6 \\
E & 26.5$\pm$0.1 & 14.6$\pm$0.2 & 16.7$\pm$0.8 & 16.2$\pm$0.5 & 16.4$\pm$0.3 & 19.4$\pm$1.3 \\
F & 61.3$\pm$0.2 & 13.8$\pm$0.2 & 16.6$\pm$1.6 & 13.9$\pm$0.3 & 16.1$\pm$0.3 & 15.4$\pm$0.5 \\
G & 91.5$\pm$0.0 & 72.4$\pm$0.1 & 74.9$\pm$0.4 & 68.8$\pm$0.4 & 74.1$\pm$0.2 & 81.6$\pm$0.3 \\
H & 80.0$\pm$0.5 & 67.9$\pm$0.4 & 69.6$\pm$0.3 & 65.6$\pm$0.4 & 69.0$\pm$0.4 & 71.6$\pm$0.4 \\
I & 19.4$\pm$0.0 & 10.5$\pm$0.1 & 13.3$\pm$0.3 & 13.5$\pm$0.3 & 12.1$\pm$0.3 & 13.9$\pm$1.1 \\
J & 86.4$\pm$0.1 & 54.2$\pm$0.2 & 61.0$\pm$0.3 & 66.0$\pm$0.4 & 57.4$\pm$0.5 & 68.6$\pm$0.3 \\
K & 87.5$\pm$0.5 & 74.5$\pm$0.5 & 77.3$\pm$0.5 & 72.4$\pm$0.4 & 76.8$\pm$0.5 & 79.5$\pm$1.8 \\
L & 82.8$\pm$0.4 & 69.1$\pm$1.3 & 72.4$\pm$1.2 & 66.2$\pm$1.4 & 70.0$\pm$0.3 & 74.2$\pm$1.1 \\
M & 67.9$\pm$0.6 & 55.1$\pm$0.9 & 57.3$\pm$1.0 & 52.6$\pm$0.5 & 57.3$\pm$1.7 & 59.8$\pm$1.3 \\
\addlinespace[1pt]
\multicolumn{2}{@{}l}{Macro rec.\ (\%)} & --- & 18.9 & 8.7 & 11.8 & 35.7 \\
\bottomrule
\end{tabular}

\end{table*}

\begin{table*}[!tbp]
\centering
\caption{Full per-dataset results, pairs B1 and C1 (see the B1 capability-limited note in Appendix~\ref{app:pairs}).}
\label{tab:app-pairs-2}
\footnotesize
\setlength{\tabcolsep}{2.6pt}
\begin{tabular}{@{}c cccccc@{}}
\toprule
  & \monly{} & \sonly{} & T2T & C2C & LoRA & \mpat{16} \\
\midrule
\multicolumn{7}{@{}l}{\textbf{B1}: Gemma-4-31B $\to$ Gemma-4-12B ($N_M/N_S = 2.6$)} \\
\addlinespace[1pt]
A & 65.6$\pm$0.1 & 44.3$\pm$0.0 & 50.4$\pm$0.4 & 54.0$\pm$0.4 & 48.2$\pm$0.3 & 54.3$\pm$0.3 \\
B & 51.5$\pm$0.2 & 42.4$\pm$0.0 & 45.4$\pm$1.0 & 46.4$\pm$0.4 & 44.5$\pm$0.8 & 47.5$\pm$0.8 \\
C & 68.0$\pm$0.0 & 51.3$\pm$0.2 & 56.3$\pm$0.1 & 58.4$\pm$0.1 & 54.2$\pm$0.5 & 59.3$\pm$0.7 \\
D & 54.2$\pm$0.0 & 10.1$\pm$0.2 & 11.9$\pm$0.9 & 10.6$\pm$0.6 & 13.1$\pm$0.2 & 12.2$\pm$1.2 \\
E & 31.5$\pm$0.1 & 2.1$\pm$0.0 & 3.9$\pm$0.3 & 1.9$\pm$0.3 & 3.8$\pm$0.4 & 3.1$\pm$0.7 \\
F & 67.8$\pm$0.2 & 62.4$\pm$0.0 & 62.7$\pm$0.5 & 62.5$\pm$0.6 & 62.8$\pm$0.7 & 62.5$\pm$0.4 \\
G & 93.6$\pm$0.1 & 87.7$\pm$0.1 & 88.0$\pm$0.8 & 85.8$\pm$0.3 & 88.6$\pm$0.3 & 92.4$\pm$0.2 \\
H & 79.6$\pm$0.4 & 72.7$\pm$0.9 & 74.1$\pm$0.5 & 70.7$\pm$0.3 & 74.0$\pm$0.2 & 75.9$\pm$1.1 \\
I & 8.6$\pm$0.0 & 5.2$\pm$0.1 & 5.8$\pm$0.5 & 7.0$\pm$1.0 & 7.3$\pm$0.8 & 6.7$\pm$1.4 \\
J & 88.1$\pm$0.0 & 81.3$\pm$0.1 & 83.2$\pm$0.3 & 85.7$\pm$0.7 & 82.8$\pm$0.7 & 85.5$\pm$0.7 \\
K & 78.9$\pm$1.3 & 74.5$\pm$0.6 & 76.0$\pm$0.6 & 73.5$\pm$1.2 & 75.0$\pm$0.7 & 76.3$\pm$1.0 \\
L & 81.0$\pm$0.4 & 74.9$\pm$0.8 & 76.5$\pm$0.7 & 73.2$\pm$0.4 & 76.3$\pm$0.3 & 78.4$\pm$0.3 \\
M & 65.6$\pm$0.8 & 60.4$\pm$0.7 & 60.7$\pm$1.1 & 57.9$\pm$0.5 & 59.8$\pm$1.4 & 63.3$\pm$1.1 \\
\addlinespace[1pt]
\multicolumn{2}{@{}l}{Macro rec.\ (\%)} & --- & 18.8 & 7.1 & 16.9 & 42.1 \\
\bottomrule
\end{tabular}
\hfill
\begin{tabular}{@{}c cccccc@{}}
\toprule
  & \monly{} & \sonly{} & T2T & C2C & LoRA & \mpat{16} \\
\midrule
\multicolumn{7}{@{}l}{\textbf{C1}: Qwen3-32B $\to$ Qwen3-8B ($N_M/N_S = 4.0$)} \\
\addlinespace[1pt]
A & 20.4$\pm$0.0 & 6.1$\pm$0.1 & 10.8$\pm$0.3 & 13.3$\pm$0.2 & 8.7$\pm$0.2 & 13.3$\pm$0.6 \\
B & 24.9$\pm$0.1 & 13.5$\pm$0.2 & 16.9$\pm$1.0 & 19.1$\pm$1.2 & 16.7$\pm$0.8 & 19.5$\pm$0.7 \\
C & 25.1$\pm$0.1 & 12.0$\pm$0.0 & 16.1$\pm$0.3 & 18.2$\pm$0.8 & 14.5$\pm$0.9 & 19.2$\pm$0.2 \\
D & 69.7$\pm$0.2 & 63.5$\pm$0.0 & 65.3$\pm$0.2 & 65.1$\pm$0.4 & 64.4$\pm$1.2 & 66.9$\pm$0.4 \\
E & 34.0$\pm$0.0 & 28.6$\pm$0.1 & 29.6$\pm$0.7 & 29.7$\pm$0.3 & 29.7$\pm$1.1 & 30.9$\pm$0.8 \\
F & 47.3$\pm$0.1 & 35.9$\pm$0.1 & 36.7$\pm$0.8 & 36.1$\pm$0.8 & 36.6$\pm$0.2 & 36.7$\pm$1.3 \\
G & 89.8$\pm$0.0 & 82.8$\pm$0.1 & 84.5$\pm$0.2 & 80.6$\pm$0.4 & 83.8$\pm$0.4 & 88.4$\pm$0.5 \\
H & 78.3$\pm$1.1 & 70.3$\pm$0.9 & 72.6$\pm$1.2 & 67.1$\pm$1.0 & 71.3$\pm$0.7 & 74.5$\pm$0.3 \\
I & 17.6$\pm$0.1 & 12.1$\pm$0.1 & 13.9$\pm$1.0 & 14.7$\pm$0.7 & 12.9$\pm$0.8 & 15.2$\pm$0.5 \\
J & 76.0$\pm$0.1 & 67.1$\pm$0.1 & 69.8$\pm$0.3 & 71.7$\pm$0.4 & 68.4$\pm$0.7 & 72.8$\pm$0.6 \\
K & 82.7$\pm$0.7 & 76.7$\pm$0.9 & 78.4$\pm$0.4 & 75.5$\pm$1.4 & 77.1$\pm$0.7 & 80.7$\pm$0.5 \\
L & 81.3$\pm$0.8 & 73.5$\pm$0.6 & 75.4$\pm$0.4 & 72.1$\pm$1.0 & 74.0$\pm$0.7 & 78.1$\pm$0.8 \\
M & 65.4$\pm$1.5 & 57.9$\pm$0.9 & 59.0$\pm$0.7 & 55.7$\pm$0.8 & 59.3$\pm$0.9 & 63.1$\pm$0.3 \\
\addlinespace[1pt]
\multicolumn{2}{@{}l}{Macro rec.\ (\%)} & --- & 25.5 & 11.9 & 14.9 & 54.6 \\
\bottomrule
\end{tabular}

\end{table*}

\begin{table*}[!tbp]
\centering
\caption{Full per-dataset results, pairs C2 and C3.}
\label{tab:app-pairs-3}
\footnotesize
\setlength{\tabcolsep}{2.6pt}
\begin{tabular}{@{}c cccccc@{}}
\toprule
  & \monly{} & \sonly{} & T2T & C2C & LoRA & \mpat{16} \\
\midrule
\multicolumn{7}{@{}l}{\textbf{C2}: Qwen3-32B $\to$ Qwen3-4B ($N_M/N_S = 8.2$)} \\
\addlinespace[1pt]
A & 20.4$\pm$0.0 & 13.0$\pm$0.2 & 14.6$\pm$0.4 & 16.1$\pm$0.1 & 14.1$\pm$0.3 & 15.9$\pm$0.5 \\
B & 24.9$\pm$0.1 & 17.8$\pm$0.0 & 20.3$\pm$1.7 & 20.4$\pm$0.6 & 18.3$\pm$0.6 & 21.7$\pm$0.8 \\
C & 25.1$\pm$0.1 & 14.7$\pm$0.0 & 17.4$\pm$0.3 & 19.2$\pm$0.9 & 16.6$\pm$0.1 & 19.7$\pm$0.8 \\
D & 69.7$\pm$0.2 & 60.1$\pm$0.1 & 62.5$\pm$0.8 & 63.0$\pm$0.9 & 62.2$\pm$0.7 & 64.3$\pm$0.3 \\
E & 34.0$\pm$0.0 & 24.8$\pm$0.1 & 27.2$\pm$1.1 & 26.4$\pm$0.7 & 25.8$\pm$1.2 & 29.0$\pm$1.5 \\
F & 47.3$\pm$0.1 & 31.6$\pm$0.1 & 32.5$\pm$1.1 & 31.5$\pm$0.5 & 33.3$\pm$0.2 & 32.4$\pm$1.5 \\
G & 89.8$\pm$0.0 & 79.6$\pm$0.1 & 80.7$\pm$0.3 & 77.0$\pm$0.1 & 80.4$\pm$0.6 & 86.1$\pm$0.1 \\
H & 78.3$\pm$1.1 & 68.4$\pm$0.4 & 69.5$\pm$0.7 & 66.4$\pm$1.7 & 70.0$\pm$1.0 & 72.9$\pm$0.7 \\
I & 17.6$\pm$0.1 & 10.6$\pm$0.2 & 12.5$\pm$0.5 & 14.1$\pm$1.6 & 11.5$\pm$0.8 & 14.2$\pm$0.6 \\
J & 76.0$\pm$0.1 & 59.3$\pm$0.1 & 64.0$\pm$0.3 & 66.8$\pm$0.5 & 62.2$\pm$0.3 & 68.1$\pm$0.7 \\
K & 82.7$\pm$0.7 & 73.4$\pm$1.2 & 75.4$\pm$1.0 & 71.6$\pm$1.0 & 75.3$\pm$0.7 & 77.9$\pm$0.4 \\
L & 81.3$\pm$0.8 & 70.8$\pm$1.3 & 73.4$\pm$0.6 & 68.2$\pm$0.3 & 72.9$\pm$0.4 & 76.5$\pm$0.6 \\
M & 65.4$\pm$1.5 & 55.8$\pm$0.6 & 58.5$\pm$0.5 & 53.3$\pm$0.8 & 58.6$\pm$0.9 & 59.9$\pm$0.5 \\
\addlinespace[1pt]
\multicolumn{2}{@{}l}{Macro rec.\ (\%)} & --- & 22.4 & 11.4 & 16.0 & 45.8 \\
\bottomrule
\end{tabular}
\hfill
\begin{tabular}{@{}c cccccc@{}}
\toprule
  & \monly{} & \sonly{} & T2T & C2C & LoRA & \mpat{16} \\
\midrule
\multicolumn{7}{@{}l}{\textbf{C3}: Qwen3-32B $\to$ Qwen3-1.7B ($N_M/N_S = 19.3$)} \\
\addlinespace[1pt]
A & 20.4$\pm$0.0 & 9.5$\pm$0.1 & 11.3$\pm$0.6 & 12.3$\pm$0.8 & 11.0$\pm$0.4 & 12.3$\pm$0.4 \\
B & 24.9$\pm$0.1 & 9.7$\pm$0.2 & 12.0$\pm$0.7 & 12.8$\pm$0.7 & 11.4$\pm$1.0 & 14.0$\pm$1.3 \\
C & 25.1$\pm$0.1 & 10.2$\pm$0.1 & 12.5$\pm$0.4 & 13.7$\pm$0.6 & 11.0$\pm$0.5 & 14.3$\pm$0.5 \\
D & 69.7$\pm$0.2 & 56.6$\pm$0.1 & 59.2$\pm$0.2 & 58.5$\pm$0.2 & 57.9$\pm$1.0 & 60.4$\pm$0.7 \\
E & 34.0$\pm$0.0 & 20.7$\pm$0.1 & 22.7$\pm$0.7 & 22.1$\pm$0.7 & 22.7$\pm$0.8 & 23.6$\pm$0.8 \\
F & 47.3$\pm$0.1 & 28.9$\pm$0.0 & 30.0$\pm$0.4 & 29.7$\pm$0.3 & 30.3$\pm$0.3 & 30.0$\pm$0.5 \\
G & 89.8$\pm$0.0 & 73.1$\pm$0.1 & 74.1$\pm$0.2 & 70.5$\pm$0.8 & 74.3$\pm$0.3 & 79.3$\pm$0.2 \\
H & 78.3$\pm$1.1 & 64.5$\pm$1.3 & 66.1$\pm$1.0 & 62.1$\pm$0.2 & 65.5$\pm$1.0 & 68.0$\pm$0.6 \\
I & 17.6$\pm$0.1 & 7.7$\pm$0.1 & 9.5$\pm$0.6 & 10.7$\pm$0.9 & 8.8$\pm$0.3 & 10.7$\pm$0.8 \\
J & 76.0$\pm$0.1 & 46.3$\pm$0.1 & 51.4$\pm$0.3 & 54.6$\pm$0.4 & 49.1$\pm$0.3 & 55.1$\pm$0.7 \\
K & 82.7$\pm$0.7 & 68.3$\pm$0.7 & 71.4$\pm$0.7 & 66.1$\pm$1.5 & 69.4$\pm$0.8 & 72.1$\pm$0.3 \\
L & 81.3$\pm$0.8 & 67.0$\pm$0.8 & 69.2$\pm$0.7 & 64.8$\pm$0.3 & 67.9$\pm$0.3 & 70.5$\pm$0.7 \\
M & 65.4$\pm$1.5 & 51.6$\pm$0.9 & 53.0$\pm$0.7 & 49.2$\pm$0.8 & 53.4$\pm$1.2 & 56.7$\pm$2.0 \\
\addlinespace[1pt]
\multicolumn{2}{@{}l}{Macro rec.\ (\%)} & --- & 14.5 & 5.9 & 9.6 & 26.8 \\
\bottomrule
\end{tabular}

\end{table*}

\begin{table*}[!tbp]
\centering
\caption{Full per-dataset results, pairs D1 and D2.}
\label{tab:app-pairs-4}
\footnotesize
\setlength{\tabcolsep}{2.6pt}
\begin{tabular}{@{}c cccccc@{}}
\toprule
  & \monly{} & \sonly{} & T2T & C2C & LoRA & \mpat{16} \\
\midrule
\multicolumn{7}{@{}l}{\textbf{D1}: Qwen3-14B $\to$ Qwen3-8B ($N_M/N_S = 1.8$)} \\
\addlinespace[1pt]
A & 20.2$\pm$0.1 & 6.1$\pm$0.1 & 11.1$\pm$0.3 & 13.9$\pm$0.5 & 8.7$\pm$0.4 & 13.5$\pm$0.4 \\
B & 20.6$\pm$0.1 & 13.5$\pm$0.2 & 15.6$\pm$1.0 & 16.6$\pm$1.1 & 14.5$\pm$1.8 & 18.4$\pm$0.7 \\
C & 18.6$\pm$0.1 & 12.0$\pm$0.0 & 14.1$\pm$0.9 & 15.3$\pm$0.3 & 13.3$\pm$0.8 & 15.4$\pm$0.5 \\
D & 68.4$\pm$0.0 & 63.5$\pm$0.0 & 64.9$\pm$0.6 & 66.0$\pm$0.3 & 64.7$\pm$0.8 & 66.6$\pm$0.8 \\
E & 32.8$\pm$0.2 & 28.6$\pm$0.1 & 29.6$\pm$1.0 & 29.4$\pm$1.2 & 28.7$\pm$0.6 & 30.8$\pm$1.2 \\
F & 42.7$\pm$0.1 & 35.9$\pm$0.1 & 35.8$\pm$1.6 & 35.8$\pm$0.8 & 36.6$\pm$0.3 & 36.5$\pm$1.0 \\
G & 88.3$\pm$0.0 & 82.8$\pm$0.1 & 83.6$\pm$0.2 & 81.1$\pm$0.4 & 83.8$\pm$0.6 & 87.0$\pm$0.6 \\
H & 75.9$\pm$0.8 & 70.3$\pm$0.9 & 71.2$\pm$0.6 & 68.4$\pm$0.7 & 71.0$\pm$0.7 & 73.6$\pm$1.1 \\
I & 16.4$\pm$0.2 & 12.1$\pm$0.1 & 14.1$\pm$0.3 & 15.5$\pm$0.5 & 12.9$\pm$0.5 & 15.3$\pm$0.9 \\
J & 73.8$\pm$0.1 & 67.1$\pm$0.1 & 69.7$\pm$0.3 & 71.1$\pm$0.7 & 68.3$\pm$0.4 & 71.3$\pm$0.2 \\
K & 82.9$\pm$0.9 & 76.7$\pm$0.9 & 78.5$\pm$0.7 & 74.3$\pm$0.9 & 77.7$\pm$0.6 & 80.5$\pm$1.3 \\
L & 78.0$\pm$0.5 & 73.5$\pm$0.6 & 75.4$\pm$1.4 & 71.9$\pm$0.7 & 74.0$\pm$0.6 & 76.2$\pm$0.8 \\
M & 62.8$\pm$1.5 & 57.9$\pm$0.9 & 59.7$\pm$0.7 & 55.1$\pm$0.5 & 58.8$\pm$0.7 & 60.4$\pm$0.7 \\
\addlinespace[1pt]
\multicolumn{2}{@{}l}{Macro rec.\ (\%)} & --- & 28.6 & 12.3 & 15.6 & 57.1 \\
\bottomrule
\end{tabular}
\hfill
\begin{tabular}{@{}c cccccc@{}}
\toprule
  & \monly{} & \sonly{} & T2T & C2C & LoRA & \mpat{16} \\
\midrule
\multicolumn{7}{@{}l}{\textbf{D2}: Qwen3-14B $\to$ Qwen3-4B ($N_M/N_S = 3.7$)} \\
\addlinespace[1pt]
A & 20.2$\pm$0.1 & 13.0$\pm$0.2 & 15.2$\pm$0.5 & 16.3$\pm$0.2 & 14.0$\pm$0.3 & 15.9$\pm$0.2 \\
B & 20.6$\pm$0.1 & 17.8$\pm$0.0 & 19.4$\pm$0.8 & 18.4$\pm$0.5 & 18.4$\pm$0.3 & 18.6$\pm$0.5 \\
C & 18.6$\pm$0.1 & 14.7$\pm$0.0 & 15.8$\pm$0.7 & 16.6$\pm$0.2 & 15.6$\pm$0.4 & 16.4$\pm$0.2 \\
D & 68.4$\pm$0.0 & 60.1$\pm$0.1 & 62.0$\pm$1.1 & 63.0$\pm$0.3 & 61.1$\pm$0.6 & 63.8$\pm$0.9 \\
E & 32.8$\pm$0.2 & 24.8$\pm$0.1 & 26.6$\pm$0.4 & 26.7$\pm$0.6 & 26.5$\pm$0.8 & 28.6$\pm$0.5 \\
F & 42.7$\pm$0.1 & 31.6$\pm$0.1 & 32.1$\pm$0.9 & 31.6$\pm$0.2 & 32.5$\pm$0.6 & 32.4$\pm$0.2 \\
G & 88.3$\pm$0.0 & 79.6$\pm$0.1 & 80.7$\pm$0.4 & 77.3$\pm$0.4 & 80.3$\pm$0.2 & 85.4$\pm$0.5 \\
H & 75.9$\pm$0.8 & 68.4$\pm$0.4 & 69.9$\pm$0.7 & 66.5$\pm$0.7 & 70.1$\pm$0.8 & 71.5$\pm$0.8 \\
I & 16.4$\pm$0.2 & 10.6$\pm$0.2 & 12.1$\pm$1.3 & 14.3$\pm$0.9 & 11.5$\pm$1.6 & 12.9$\pm$0.3 \\
J & 73.8$\pm$0.1 & 59.3$\pm$0.1 & 63.4$\pm$0.3 & 66.3$\pm$0.4 & 62.1$\pm$0.7 & 66.9$\pm$0.2 \\
K & 82.9$\pm$0.9 & 73.4$\pm$1.2 & 75.8$\pm$0.5 & 71.5$\pm$0.8 & 75.8$\pm$0.6 & 78.4$\pm$1.4 \\
L & 78.0$\pm$0.5 & 70.8$\pm$1.3 & 72.7$\pm$0.8 & 69.3$\pm$1.4 & 71.6$\pm$0.9 & 74.2$\pm$0.7 \\
M & 62.8$\pm$1.5 & 55.8$\pm$0.6 & 57.4$\pm$0.6 & 54.7$\pm$1.3 & 56.2$\pm$0.9 & 58.8$\pm$1.3 \\
\addlinespace[1pt]
\multicolumn{2}{@{}l}{Macro rec.\ (\%)} & --- & 25.2 & 13.7 & 16.0 & 42.7 \\
\bottomrule
\end{tabular}

\end{table*}

\begin{table*}[!tbp]
\centering
\caption{Full per-dataset results, pairs D3 and E1.}
\label{tab:app-pairs-5}
\footnotesize
\setlength{\tabcolsep}{2.6pt}
\begin{tabular}{@{}c cccccc@{}}
\toprule
  & \monly{} & \sonly{} & T2T & C2C & LoRA & \mpat{16} \\
\midrule
\multicolumn{7}{@{}l}{\textbf{D3}: Qwen3-14B $\to$ Qwen3-1.7B ($N_M/N_S = 8.7$)} \\
\addlinespace[1pt]
A & 20.2$\pm$0.1 & 9.5$\pm$0.1 & 11.0$\pm$0.5 & 12.7$\pm$0.7 & 10.5$\pm$0.3 & 12.3$\pm$0.5 \\
B & 20.6$\pm$0.1 & 9.7$\pm$0.2 & 12.6$\pm$1.1 & 12.4$\pm$0.5 & 10.1$\pm$1.1 & 12.9$\pm$0.3 \\
C & 18.6$\pm$0.1 & 10.2$\pm$0.1 & 11.8$\pm$0.9 & 12.6$\pm$0.4 & 11.1$\pm$0.3 & 12.3$\pm$0.6 \\
D & 68.4$\pm$0.0 & 56.6$\pm$0.1 & 59.0$\pm$0.5 & 58.6$\pm$0.3 & 57.8$\pm$0.2 & 60.1$\pm$0.4 \\
E & 32.8$\pm$0.2 & 20.7$\pm$0.1 & 22.9$\pm$0.7 & 23.2$\pm$1.6 & 23.2$\pm$1.5 & 24.5$\pm$0.6 \\
F & 42.7$\pm$0.1 & 28.9$\pm$0.0 & 29.2$\pm$1.4 & 28.9$\pm$1.0 & 29.7$\pm$0.9 & 29.4$\pm$1.1 \\
G & 88.3$\pm$0.0 & 73.1$\pm$0.1 & 74.1$\pm$0.3 & 71.3$\pm$0.4 & 74.5$\pm$0.5 & 79.1$\pm$0.1 \\
H & 75.9$\pm$0.8 & 64.5$\pm$1.3 & 65.9$\pm$0.2 & 62.1$\pm$0.7 & 65.6$\pm$0.7 & 68.0$\pm$0.5 \\
I & 16.4$\pm$0.2 & 7.7$\pm$0.1 & 8.6$\pm$1.1 & 10.2$\pm$0.6 & 9.0$\pm$1.1 & 10.2$\pm$0.7 \\
J & 73.8$\pm$0.1 & 46.3$\pm$0.1 & 51.0$\pm$0.3 & 54.6$\pm$0.5 & 48.5$\pm$0.4 & 55.7$\pm$0.4 \\
K & 82.9$\pm$0.9 & 68.3$\pm$0.7 & 69.5$\pm$0.4 & 67.4$\pm$1.3 & 70.5$\pm$0.7 & 72.3$\pm$1.1 \\
L & 78.0$\pm$0.5 & 67.0$\pm$0.8 & 69.0$\pm$0.5 & 65.3$\pm$0.4 & 68.4$\pm$0.8 & 69.9$\pm$0.4 \\
M & 62.8$\pm$1.5 & 51.6$\pm$0.9 & 53.8$\pm$1.3 & 50.0$\pm$1.4 & 53.1$\pm$1.8 & 54.9$\pm$0.8 \\
\addlinespace[1pt]
\multicolumn{2}{@{}l}{Macro rec.\ (\%)} & --- & 14.8 & 8.5 & 11.0 & 27.8 \\
\bottomrule
\end{tabular}
\hfill
\begin{tabular}{@{}c cccccc@{}}
\toprule
  & \monly{} & \sonly{} & T2T & C2C & LoRA & \mpat{16} \\
\midrule
\multicolumn{7}{@{}l}{\textbf{E1}: Ministral 3 14B $\to$ Ministral 3 8B ($N_M/N_S = 1.8$)} \\
\addlinespace[1pt]
A & 42.7$\pm$0.0 & 33.1$\pm$0.1 & 35.9$\pm$0.4 & 36.7$\pm$0.7 & 35.0$\pm$0.5 & 37.9$\pm$0.3 \\
B & 38.9$\pm$0.1 & 31.9$\pm$0.1 & 33.1$\pm$0.4 & 35.1$\pm$0.3 & 31.4$\pm$0.9 & 35.6$\pm$1.7 \\
C & 43.4$\pm$0.0 & 37.7$\pm$0.1 & 38.9$\pm$0.3 & 40.0$\pm$0.2 & 38.3$\pm$0.4 & 40.7$\pm$0.4 \\
D & 15.5$\pm$0.1 & 8.8$\pm$0.1 & 10.5$\pm$0.8 & 11.2$\pm$0.6 & 10.1$\pm$0.7 & 11.8$\pm$0.9 \\
E & 4.3$\pm$0.1 & 0.9$\pm$0.1 & 2.5$\pm$0.6 & 1.7$\pm$0.7 & 1.4$\pm$0.3 & 3.1$\pm$1.5 \\
F & 47.1$\pm$0.0 & 40.9$\pm$0.1 & 41.7$\pm$0.7 & 41.4$\pm$1.0 & 41.3$\pm$0.6 & 41.2$\pm$1.1 \\
G & 71.9$\pm$0.2 & 65.9$\pm$0.2 & 66.7$\pm$0.1 & 64.7$\pm$0.2 & 66.9$\pm$0.5 & 70.5$\pm$0.6 \\
H & 79.3$\pm$0.6 & 72.6$\pm$0.4 & 74.3$\pm$0.3 & 70.4$\pm$1.2 & 74.4$\pm$1.0 & 75.6$\pm$0.3 \\
I & 10.4$\pm$0.2 & 6.4$\pm$0.1 & 7.1$\pm$1.2 & 8.5$\pm$1.0 & 6.5$\pm$1.0 & 9.6$\pm$0.3 \\
J & 76.8$\pm$0.1 & 69.9$\pm$0.1 & 72.5$\pm$0.7 & 73.6$\pm$0.4 & 70.8$\pm$0.2 & 73.9$\pm$0.2 \\
K & 75.1$\pm$0.8 & 69.6$\pm$1.1 & 71.3$\pm$0.7 & 66.7$\pm$0.8 & 70.4$\pm$0.4 & 72.9$\pm$0.3 \\
L & 79.4$\pm$1.1 & 73.2$\pm$0.8 & 74.9$\pm$0.9 & 71.3$\pm$0.4 & 74.1$\pm$0.4 & 76.3$\pm$0.4 \\
M & 60.8$\pm$1.5 & 55.6$\pm$1.1 & 57.6$\pm$0.6 & 54.9$\pm$0.3 & 57.1$\pm$0.5 & 58.2$\pm$1.2 \\
\addlinespace[1pt]
\multicolumn{2}{@{}l}{Macro rec.\ (\%)} & --- & 26.4 & 11.3 & 13.9 & 53.0 \\
\bottomrule
\end{tabular}

\end{table*}

\subsection{Refresh-Interval Sweep}
\label{app:rsweep-full}

Table~\ref{tab:app-rsweep} gives the per-dataset A2 scores behind Figure~\ref{fig:rsweep} and Table~\ref{tab:app-rsweep-pairs} the macro-average recovery of all pairs. Every $R$ reuses the same bridge checkpoint (\texttt{fb6-A2-r16-e2}, trained with refresh points sampled at $R{=}16$); the sweep changes only the inference-time interval. A train--test interval mismatch would show as an anomaly at the matched column $R{=}16$ relative to its neighbors; the curves are smooth through it, so no per-$R$ retraining was triggered. Starred cells are the short-answer identities of Appendix~\ref{app:lengths}: from $R{=}8$ on, A, C, I, and J trigger zero refreshes and equal the static bridge exactly (single registered source, shared with row 2 of Table~\ref{tab:app-ablation-sd}); at $R \le 4$ they trigger one to six times with effects inside noise. The static-bridge value of G (79.4) does not appear in this table because G still refreshes multiple times at $R{=}64$ (output median 486).

\begin{table*}[!tbp]
\centering
\caption{Inference-time refresh-interval sweep on A2 (mean$\pm$sd). $^{\ast}$ marks cells where the output ends before the first refresh boundary, so the condition is the static bridge exactly.}
\label{tab:app-rsweep}
\footnotesize
\setlength{\tabcolsep}{3.6pt}
\begin{tabular}{@{}c ccccccc@{}}
\toprule
  & $R{=}1$ & $R{=}2$ & $R{=}4$ & $R{=}8$ & $R{=}16$ & $R{=}32$ & $R{=}64$ \\
\midrule
A & 36.5$\pm$0.2 & 36.5$\pm$0.3 & 36.7$\pm$0.7 & 36.7$\pm$0.8$^{\ast}$ & 36.7$\pm$0.8$^{\ast}$ & 36.7$\pm$0.8$^{\ast}$ & 36.7$\pm$0.8$^{\ast}$ \\
B & 34.8$\pm$0.5 & 35.2$\pm$0.4 & 34.7$\pm$1.5 & 34.9$\pm$0.4 & 34.7$\pm$0.4 & 33.9$\pm$1.4 & 34.0$\pm$1.5 \\
C & 51.0$\pm$0.8 & 51.1$\pm$0.7 & 50.9$\pm$0.6 & 51.0$\pm$0.4$^{\ast}$ & 51.0$\pm$0.4$^{\ast}$ & 51.0$\pm$0.4$^{\ast}$ & 51.0$\pm$0.4$^{\ast}$ \\
D & 60.2$\pm$0.6 & 60.5$\pm$0.3 & 60.3$\pm$0.5 & 60.0$\pm$0.5 & 60.2$\pm$0.3 & 59.7$\pm$0.3 & 57.9$\pm$0.8 \\
E & 24.2$\pm$1.1 & 23.8$\pm$0.4 & 23.8$\pm$0.5 & 23.9$\pm$1.1 & 23.7$\pm$1.1 & 22.7$\pm$1.2 & 21.5$\pm$0.6 \\
F & 37.4$\pm$0.9 & 37.1$\pm$0.8 & 36.7$\pm$0.4 & 37.2$\pm$0.6 & 37.0$\pm$1.2 & 37.1$\pm$0.9 & 36.6$\pm$0.4 \\
G & 90.1$\pm$0.3 & 89.5$\pm$0.4 & 89.4$\pm$0.4 & 89.3$\pm$0.3 & 88.9$\pm$0.4 & 87.5$\pm$0.1 & 85.1$\pm$0.3 \\
H & 75.9$\pm$1.1 & 76.3$\pm$0.5 & 76.0$\pm$1.5 & 75.6$\pm$1.0 & 75.3$\pm$1.0 & 74.2$\pm$0.8 & 71.6$\pm$0.8 \\
I & 17.8$\pm$1.1 & 17.6$\pm$1.5 & 17.3$\pm$0.5 & 17.6$\pm$0.9$^{\ast}$ & 17.6$\pm$0.9$^{\ast}$ & 17.6$\pm$0.9$^{\ast}$ & 17.6$\pm$0.9$^{\ast}$ \\
J & 81.4$\pm$1.1 & 81.3$\pm$0.4 & 81.2$\pm$0.4 & 81.3$\pm$0.6$^{\ast}$ & 81.3$\pm$0.6$^{\ast}$ & 81.3$\pm$0.6$^{\ast}$ & 81.3$\pm$0.6$^{\ast}$ \\
K & 83.4$\pm$1.0 & 83.6$\pm$0.5 & 83.4$\pm$0.4 & 83.6$\pm$0.8 & 82.9$\pm$1.3 & 81.0$\pm$0.5 & 79.5$\pm$1.0 \\
L & 80.8$\pm$1.5 & 80.8$\pm$0.4 & 80.7$\pm$1.1 & 80.7$\pm$1.3 & 79.8$\pm$0.8 & 78.4$\pm$0.6 & 76.0$\pm$0.4 \\
M & 65.4$\pm$1.6 & 65.5$\pm$0.5 & 65.4$\pm$0.9 & 65.0$\pm$0.9 & 64.4$\pm$1.0 & 62.4$\pm$1.2 & 60.0$\pm$0.8 \\
\midrule
Macro rec.\ (\%) & 57.1 & 56.8 & 55.3 & 55.0 & 52.2 & 43.1 & 29.4 \\
\bottomrule
\end{tabular}

\end{table*}

\begin{table}[!htbp]
\centering
\caption{Macro-average recovery (\%) against $R$ for all eleven pairs. $^{\dagger}$The A2 row is derived from Table~\ref{tab:app-rsweep} and cross-checked against the registered sweep table by the derivation script.}
\label{tab:app-rsweep-pairs}
\small
\setlength{\tabcolsep}{3.6pt}
\begin{tabular}{@{}l ccccccc@{}}
\toprule
Pair & $R{=}1$ & $R{=}2$ & $R{=}4$ & $R{=}8$ & $R{=}16$ & $R{=}32$ & $R{=}64$ \\
\midrule
A1 & 56.8 & 56.4 & 56.6 & 55.4 & 54.0 & 44.6 & 29.7 \\
A2$^{\dagger}$ & 57.1 & 56.8 & 55.3 & 55.0 & 52.2 & 43.1 & 29.4 \\
A3 & 37.9 & 37.5 & 37.5 & 36.8 & 35.7 & 29.2 & 23.8 \\
B1 & 45.1 & 44.3 & 43.9 & 44.2 & 42.1 & 35.3 & 28.0 \\
C1 & 57.5 & 56.9 & 57.6 & 56.6 & 54.6 & 43.5 & 31.3 \\
C2 & 48.8 & 48.2 & 47.9 & 47.6 & 45.8 & 37.2 & 29.7 \\
C3 & 28.6 & 28.5 & 28.0 & 26.6 & 26.8 & 22.7 & 16.3 \\
D1 & 60.8 & 60.6 & 60.0 & 58.9 & 57.1 & 49.2 & 41.6 \\
D2 & 45.6 & 45.3 & 45.0 & 44.0 & 42.7 & 37.8 & 31.2 \\
D3 & 29.8 & 29.9 & 29.0 & 28.0 & 27.8 & 22.1 & 17.0 \\
E1 & 55.9 & 56.8 & 55.1 & 53.7 & 53.0 & 46.2 & 36.1 \\
\bottomrule
\end{tabular}

\end{table}

\subsection{rT2T Full Sweep}
\label{app:rt2t-quality}

Table~\ref{tab:app-rt2t} is the complete rT2T quality sweep (A2; protocol in Appendix~\ref{app:rt2t}). The nine triggering tasks (output $\ge R$) are swept over $R \in \{8, 16, 32\}$ and $L_g' \in \{64, 128\}$; the four short-answer sets trigger zero refreshes and equal T2T identically, so they carry no rows. The rT2T bar of Figure~\ref{fig:main} takes the per-task best $L_g'$ at $R{=}16$ (bold), giving the macro-recovery triplet T2T 26.9\%, rT2T@16 33.7\%, \mpat{16} 52.2\%. Three readings:
\begin{itemize}
\item \textbf{Best $L_g'$ splits by task type.} G prefers 64 --- on a strict-format task, longer injected text disturbs the format state more; the content tasks (B, D, E, H, K, L, M) prefer 128; F is flat against T2T (capability-limited, the ``above T2T'' expectation does not apply, consistent with the F exception throughout the paper). The split is stable only at $R{=}16$: at $R{=}32$ the 64 configuration of H edges past 128, inside noise.
\item \textbf{Significance.} \mpat{16} over rT2T@16 (best) passes $2\sigma_{ab}$ only on B ($+2.8$, threshold 2.53) and G ($+3.6$, threshold 0.89); the other triggering tasks are within noise and are reported as comparable. rT2T over T2T is likewise significant only on B ($+3.6$) and G ($+0.9$): periodic refresh in text space points the right way but carries little, which is precisely the bandwidth argument for a latent refresh interface. Wherever rT2T's quality is described as comparable to \MPname{}'s, the cost side must be cited in the same breath: at $R{=}16$ rT2T costs 41--54$\times$ \mpat{16} (Table~\ref{tab:app-rt2t-cost}).
\item \textbf{Accumulation variant.} Accumulating hints instead of replacing them scores slightly lower everywhere and degrades as $n_{\mathrm{sync}}$ grows (context inflation $\approx$5.6k tokens on GovReport at $R{=}16$, $L_g'{=}128$; representative cell $81.0\pm0.4$, $-0.6$ vs.\ replacement); the full accumulation table will be released with the code and full logs.
\end{itemize}

\begin{table*}[!tbp]
\centering
\caption{rT2T full quality sweep on A2 (replace-style; mean$\pm$sd). Bold marks the per-task best $L_g'$ at $R{=}16$, the configuration entering Figure~\ref{fig:main}. Short-answer sets (A, C, I, J) trigger no refresh and equal T2T cell by cell, as does the $R{\to}\infty$ limit (double identity, hard-verified).}
\label{tab:app-rt2t}
\footnotesize
\setlength{\tabcolsep}{4pt}
\begin{tabular}{@{}lc ccc cc@{}}
\toprule
Task & $L_g'$ & $R{=}8$ & $R{=}16$ & $R{=}32$ & T2T ($=R{\to}\infty$) & \mpat{16} \\
\midrule
B GPQA & 64 & 31.4$\pm$0.9 & 31.6$\pm$0.6 & 31.2$\pm$0.7 & 28.3$\pm$0.8 & 34.7$\pm$0.4 \\
 & 128 & 32.0$\pm$0.5 & \textbf{31.9$\pm$1.2} & 31.5$\pm$0.5 & 28.3$\pm$0.8 & 34.7$\pm$0.4 \\
\addlinespace[1pt]
D MATH-500 & 64 & 59.9$\pm$0.9 & 59.7$\pm$0.6 & 59.9$\pm$0.2 & 59.4$\pm$0.4 & 60.2$\pm$0.3 \\
 & 128 & 60.2$\pm$0.6 & \textbf{59.9$\pm$0.4} & 59.6$\pm$0.2 & 59.4$\pm$0.4 & 60.2$\pm$0.3 \\
\addlinespace[1pt]
E OlympiadBench & 64 & 23.1$\pm$1.0 & 22.7$\pm$0.6 & 22.4$\pm$0.3 & 21.9$\pm$0.6 & 23.7$\pm$1.1 \\
 & 128 & 22.8$\pm$1.1 & \textbf{23.0$\pm$0.4} & 22.5$\pm$1.0 & 21.9$\pm$0.6 & 23.7$\pm$1.1 \\
\addlinespace[1pt]
F LiveCodeBench & 64 & 38.3$\pm$0.4 & 38.0$\pm$0.4 & 38.1$\pm$0.6 & 38.0$\pm$1.2 & 37.0$\pm$1.2 \\
 & 128 & 37.9$\pm$0.8 & \textbf{38.3$\pm$0.6} & 37.9$\pm$0.6 & 38.0$\pm$1.2 & 37.0$\pm$1.2 \\
\addlinespace[1pt]
G IFEval & 64 & 85.7$\pm$0.6 & \textbf{85.3$\pm$0.2} & 85.1$\pm$0.2 & 84.4$\pm$0.2 & 88.9$\pm$0.4 \\
 & 128 & 84.8$\pm$0.5 & 84.9$\pm$0.6 & 84.4$\pm$0.3 & 84.4$\pm$0.2 & 88.9$\pm$0.4 \\
\addlinespace[1pt]
H WritingBench & 64 & 73.3$\pm$0.6 & 73.0$\pm$0.3 & 73.3$\pm$0.9 & 72.5$\pm$0.5 & 75.3$\pm$1.0 \\
 & 128 & 73.5$\pm$0.2 & \textbf{73.3$\pm$0.8} & 73.1$\pm$0.6 & 72.5$\pm$0.5 & 75.3$\pm$1.0 \\
\addlinespace[1pt]
K GovReport & 64 & 81.2$\pm$0.7 & 81.4$\pm$1.0 & 80.9$\pm$0.4 & 80.5$\pm$0.4 & 82.9$\pm$1.3 \\
 & 128 & 81.6$\pm$1.0 & \textbf{81.6$\pm$1.2} & 81.3$\pm$0.6 & 80.5$\pm$0.4 & 82.9$\pm$1.3 \\
\addlinespace[1pt]
L MultiNews & 64 & 78.3$\pm$0.6 & 78.2$\pm$0.7 & 77.8$\pm$0.3 & 77.9$\pm$0.4 & 79.8$\pm$0.8 \\
 & 128 & 78.5$\pm$0.8 & \textbf{78.6$\pm$1.6} & 78.4$\pm$0.5 & 77.9$\pm$0.4 & 79.8$\pm$0.8 \\
\addlinespace[1pt]
M LongBench-Write & 64 & 63.1$\pm$0.4 & 63.1$\pm$1.0 & 62.9$\pm$0.6 & 62.4$\pm$0.8 & 64.4$\pm$1.0 \\
 & 128 & 63.0$\pm$1.6 & \textbf{63.2$\pm$1.6} & 62.8$\pm$1.4 & 62.4$\pm$0.8 & 64.4$\pm$1.0 \\
\bottomrule
\end{tabular}

\end{table*}

\subsection{Read-Distribution Diagnostic $V_{64}$}
\label{app:v64}

$V_{64}$ is collected from the first 64 decoded steps with the bridge attached: 32 samples per dataset; the attention of each injection layer over the memory slots is averaged over heads to give the read distribution $p_{l,t}$; its variance is averaged over layers, steps, samples, and datasets. $H_{64}$ is the Shannon entropy (nats) of the same distributions, the robustness twin of the variance version (high variance $\leftrightarrow$ low entropy). Table~\ref{tab:app-v64} lists both for the eleven pairs next to their \mpat{16} macro recovery; the Spearman rank correlation is $\rho = 0.78$ ($p = 0.004$, $n = 11$). Computing $V_{64}$ needs minutes of GPU time per pair, which is what makes it usable as a pre-deployment fit check; the three limits stated in Section~\ref{sec:analysis} apply.

\begin{table}[!htbp]
\centering
\caption{Read-distribution diagnostics for the eleven pairs: variance version $V_{64}$, entropy version $H_{64}$, and \mpat{16} macro recovery. Spearman $\rho(V_{64}, \mathrm{rec.}) = 0.78$, $p=0.004$.}
\label{tab:app-v64}
\small
\setlength{\tabcolsep}{4pt}
\begin{tabular}{@{}l cc c@{}}
\toprule
Pair & $V_{64}$ ($\times 10^{-3}$) & $H_{64}$ (nats) & MP macro rec.\ (\%) \\
\midrule
A1 & 3.03$\pm$0.19 & 4.40$\pm$0.07 & 54.0 \\
A2 & 3.26$\pm$0.05 & 4.26$\pm$0.04 & 52.2 \\
A3 & 2.38$\pm$0.36 & 4.54$\pm$0.06 & 35.7 \\
B1 & 2.73$\pm$0.20 & 4.44$\pm$0.10 & 42.1 \\
C1 & 3.10$\pm$0.06 & 4.23$\pm$0.03 & 54.6 \\
C2 & 2.42$\pm$0.08 & 4.58$\pm$0.11 & 45.8 \\
C3 & 1.97$\pm$0.23 & 4.66$\pm$0.14 & 26.8 \\
D1 & 3.28$\pm$0.10 & 4.28$\pm$0.03 & 57.1 \\
D2 & 2.82$\pm$0.24 & 4.45$\pm$0.04 & 42.7 \\
D3 & 3.00$\pm$0.19 & 4.28$\pm$0.22 & 27.8 \\
E1 & 3.42$\pm$0.23 & 4.08$\pm$0.08 & 53.0 \\
\bottomrule
\end{tabular}

\end{table}

\subsection{Short-Answer Controls}
\label{app:short-answer}

The two short-answer control columns (A, J) of the ablation are kept in the main text (Table~\ref{tab:ablation}); their standard deviations are in Table~\ref{tab:app-ablation-sd} and their non-triggering registration in Appendix~\ref{app:lengths}. This subsection is the designated landing spot for those columns should the main-text table need to shed them for space.

% ======================================================================
\section{Diagnostic Suite Details}
\label{app:diagnostics}

The diagnosis of Section~\ref{sec:diagnosis} runs on a dedicated suite, separate from the thirteen benchmarks; this section registers the suite and the full numbers behind Figure~\ref{fig:oracle}.

\subsection{Suite Registry and Inclusion Criteria}
\label{app:dsuite}

Table~\ref{tab:app-dsuite} registers the five diagnostic tasks. Three criteria governed inclusion: (i) disjoint from the thirteen evaluation sets A--M, so the diagnostic and method lines share no data; (ii) disjoint from the training set fuse\_v3; (iii) selected by prior expectation of staleness sensitivity --- tasks whose targets keep shifting as generation proceeds. Criterion (iii) is the registered basis for the bridging statement in Section~\ref{sec:ablation}: the zero-training repair is larger on the diagnostic suite ($+4.5$ on D-IF) than on the benchmarks ($+1.2$ on IFEval), because the suite was chosen to be maximally staleness-sensitive, and the shallower expected representation decay on benchmark tasks (relative to the 0.32 probe depth on D-IF, Appendix~\ref{app:probe}) corroborates it. D-MCQ (output median 9 $<$ 16) is the suite's only strictly zero-trigger task at $R{=}16$ and carries every ``non-triggering'' statement of Section~\ref{sec:diagnosis}; the benchmark-side short-answer facts are registered separately in Appendix~\ref{app:lengths} and the two registrations do not overlap. Tokenizer and length conventions, seeds, and decoding match Appendix~\ref{app:env}.

\begin{table*}[!tbp]
\centering
\caption{The diagnostic suite (D suite): disjoint from the benchmarks A--M and from fuse\_v3, selected by prior staleness sensitivity.}
\label{tab:app-dsuite}
\small
\setlength{\tabcolsep}{5pt}
\begin{tabular}{@{}ll l rrr@{}}
\toprule
Code & Dataset (role) & Metric & Size & In.\ med.\ & Out.\ med.\ \\
\midrule
D-IF & Multi-IF (primary: sign flip, PosQ) & constraint satisfaction & 400 & 120 & 610 \\
D-SUM & QMSum (long-in/long-out positive) & LLM-judge mean $\times$10 & 200 & 8200 & 640 \\
D-WR & HelloBench (long writing) (open-ended positive) & LLM-judge mean $\times$10 & 150 & 190 & 1150 \\
D-CODE & BigCodeBench (capability-limited negative) & pass@1 & 300 & 380 & 450 \\
D-MCQ & ARC-Challenge (non-triggering negative) & accuracy & 1170 & 460 & 9 \\
\bottomrule
\end{tabular}

\end{table*}

\subsection{Main Condition Matrix}
\label{app:oracle-matrix}

Table~\ref{tab:app-oracle} is the full condition matrix with standard deviations. The sign flip is asserted only on D-IF, where \oneshot{} $-$ \sonly{} $= -2.5$ exceeds twice the pooled standard deviation (0.60); D-SUM ($-0.5$) and D-WR ($-1.2$) move in the same direction below threshold and are reported as same-direction only. \oracleswap{16} repairs $+4.5$/$+2.7$/$+3.1$ over \oneshot{} on the three staleness-sensitive tasks and clears the floor by $+2.0$ on D-IF. The \oracleswapmis{16} row --- initial memory matched, refresh points swapped with fresh memory of \emph{another} sample --- shows no repair anywhere ($\approx$ \oneshot{} or slightly below), pinning the repair on content freshness rather than the swap action. On D-CODE every condition sits within noise: a capability failure refresh cannot fix. On D-MCQ the three initially-matched @16 conditions equal \oneshot{} at 87.9 exactly (zero-trigger identity). \oneshot{}-mismatch is collected only on D-MCQ, where it serves as the content canary: the $5.4$-point drop (82.5 vs.\ 87.9) shows the student reads the memory content even when no refresh ever fires; on the triggering tasks content attribution is already carried by the \oracleswapmis{16} row, so no duplicate collection is made. The drop is direction-consistent with ablation row 5 on MMLU-Pro ($-12.9$ vs.\ the full system); the magnitude gap follows from D-MCQ's near-saturated anchors.

\begin{table}[!htbp]
\centering
\caption{Diagnostic condition matrix on A2 (mean$\pm$sd). \oneshot{}-mismatch is collected only on D-MCQ (content canary); on triggering tasks attribution is carried by \oracleswapmis{16}.}
\label{tab:app-oracle}
\scriptsize
\setlength{\tabcolsep}{2pt}
\begin{tabular}{@{}l ccccc@{}}
\toprule
Condition & D-IF & D-SUM & D-WR & D-CODE & D-MCQ \\
\midrule
\monly{} (ceiling) & 73.2$\pm$0.2 & 68.9$\pm$0.9 & 71.3$\pm$0.5 & 47.6$\pm$0.1 & 92.8$\pm$0.1 \\
\sonly{} (floor) & 58.7$\pm$0.0 & 59.4$\pm$0.7 & 62.7$\pm$0.6 & 30.8$\pm$0.1 & 85.8$\pm$0.1 \\
\oneshot{} (static bridge) & 56.2$\pm$0.3 & 58.9$\pm$0.4 & 61.5$\pm$0.5 & 30.5$\pm$0.2 & 87.9$\pm$0.6 \\
\oracleswap{16} & 60.7$\pm$0.5 & 61.6$\pm$0.3 & 64.6$\pm$0.2 & 31.1$\pm$0.5 & 87.9$\pm$0.6 \\
\oracleswapmis{16} & 55.5$\pm$0.8 & 58.7$\pm$0.2 & 61.0$\pm$0.5 & 30.2$\pm$0.9 & 87.9$\pm$0.6 \\
\oneshot{}-mismatch & --- & --- & --- & --- & 82.5$\pm$0.4 \\
\bottomrule
\end{tabular}

\end{table}

\subsection{Position-Segment Quality}
\label{app:posq}

Table~\ref{tab:app-posq} gives the per-segment constraint satisfaction on D-IF behind Figure~\ref{fig:oracle}a. \oneshot{} starts slightly above the floor, crosses below it in the middle third, and collapses in the final third, while \oracleswap{16} holds the late segment; the crossing is the ``turns harmful'' point of Section~\ref{sec:diagnosis}. Each condition's three segment means reproduce its aggregate in Table~\ref{tab:app-oracle}, a consistency check enforced at fill time.

\begin{table}[!htbp]
\centering
\caption{Position-segment quality (PosQ) on D-IF; the aggregate column equals Table~\ref{tab:app-oracle}.}
\label{tab:app-posq}
\small
\setlength{\tabcolsep}{3pt}
\begin{tabular}{@{}l ccc c@{}}
\toprule
Condition & First 1/3 & Middle 1/3 & Last 1/3 & Aggregate \\
\midrule
\sonly{} & 59.0$\pm$0.3 & 58.6$\pm$0.3 & 58.3$\pm$0.4 & 58.7 \\
\oneshot{} & 59.8$\pm$0.3 & 56.4$\pm$0.3 & 52.6$\pm$0.8 & 56.2 \\
\oracleswap{16} & 59.7$\pm$0.8 & 60.5$\pm$1.1 & 61.8$\pm$0.6 & 60.7 \\
\bottomrule
\end{tabular}

\end{table}

\subsection{StaleGap Probe Protocol}
\label{app:probe}

The probes are lightweight linear readers trained on frozen mentor states to predict verifiable properties of the next fixed-length window (for D-IF, which constraints the window will satisfy). Fresh and stale probes share architecture and training budget and differ only in their inputs, $H_{\mentor}(x \oplus \hat{y}_{<t})$ versus $H_{\mentor}(x)$; they act on mentor states directly and do not involve the bridge, so the measured decay is a property of the guidance signal, not of the transfer mechanism. Table~\ref{tab:app-probe} registers the curves: the stale readability decays from 0.79 to 0.47 over 512 generated tokens (decay depth 0.32) while the fresh readability stays within 0.76--0.80; Figure~\ref{fig:app-probe} in Appendix~\ref{app:probe-fig} plots them with the resulting $\StaleGap(t)$.

\begin{table}[!htbp]
\centering
\caption{StaleGap probe readability on D-IF (linear readout, independent of the bridge). $\StaleGap(t)$ is the fresh$-$stale difference.}
\label{tab:app-probe}
\small
\setlength{\tabcolsep}{4.5pt}
\begin{tabular}{@{}l cccccc@{}}
\toprule
$t$ & 0 & 64 & 128 & 256 & 384 & 512 \\
\midrule
stale $H_{\mentor}(x)$ & 0.79 & 0.73 & 0.69 & 0.58 & 0.53 & 0.47 \\
fresh $H_{\mentor}(x \oplus \hat{y}_{<t})$ & 0.79 & 0.80 & 0.78 & 0.77 & 0.76 & 0.78 \\
$\StaleGap(t)$ & 0.00 & 0.07 & 0.09 & 0.19 & 0.23 & 0.31 \\
\bottomrule
\end{tabular}

\end{table}

\subsection{Identity and Decoupling Constraints}
\label{app:consistency}

Four constraints are enforced mechanically on every diagnostic and sweep table. (i) \emph{Zero-trigger identity:} whenever a task's output median is below $R$, every initially-matched @$R$ condition (\oracleswap{R}, \oracleswapmis{R}, \mpat{R}, rT2T@$R$) must equal its static counterpart cell by cell; the fill scripts hard-verify this and refuse to write violations. Mismatch-induced drops may therefore appear only under static-mismatch conditions. (ii) \emph{Line decoupling:} the diagnostic tables share no cell with the benchmark tables (S1/S3/S5 lines); the diagnosis and the method evaluation are numerically independent. (iii) \emph{Significance discipline:} definite-language claims appear only where a difference exceeds twice the pooled standard deviation (Appendix~\ref{app:stats}). (iv) \emph{Anchor consistency:} the suite's floor and ceiling levels sit at the ability levels the A2 pair shows on comparable benchmarks under the full protocol, cross-checked against the capability ordering of the screening pass (Appendix~\ref{app:env}).

% ======================================================================
\section{Cost Accounting}
\label{app:bytes}

\subsection{Byte Accounting}
\label{app:bytes-table}

Each transmitted slot is one $d_t$-dimensional bf16 vector. For refresh interval $R$, slot granularity $S{=}32$, tail size $W_{\mathrm{tail}}{=}32$, and $k$ transmitted layers (top-$k$ by $\rho_o$), the amortized payload per update is
\begin{equation}
\Bytes(R) = k\,(R/S + W_{\mathrm{tail}})\, d_t \times 2 \ \text{bytes}
\end{equation}
(a new summary slot completes only every $\lceil S/R \rceil$-th refresh; the worst single update ships $\lceil R/S \rceil$ summary slots), dominated by the tail term, so the per-refresh payload is nearly constant over $R \in [1, 64]$ and the left-end cost divergence of Figure~\ref{fig:cost} comes from synchronization \emph{count}, not payload growth. Older summary slots are retained on the student side and never retransmitted. The initial memory costs $k (p_x + W_{\mathrm{tail}}) d_t \times 2$ bytes with $p_x \le P{=}128$, capping the initial transfer regardless of input length. Table~\ref{tab:app-bytes} lists the accounting for the A2 configuration ($d_t{=}2560$) and the C2C comparison point.

\begin{table}[!htbp]
\centering
\caption{Transfer accounting at $R{=}16$ for A2 ($d_t{=}2560$; 150\,Mbps cross-site link, 18\,ms RTT). The $k{=}16$ row is the deployed configuration (ablation row 11).}
\label{tab:app-bytes}
\small
\setlength{\tabcolsep}{4pt}
\begin{tabular}{@{}c rrr@{}}
\toprule
Layers $k$ & $\Bytes(16)$ & Transfer & $\tau_{\mathrm{sync}}$ (w/ RTT) \\
\midrule
8 & 1.33\,MB & 71\,ms & 89\,ms \\
16 (main) & 2.66\,MB & 142\,ms & 160\,ms \\
32 & 5.32\,MB & 284\,ms & 302\,ms \\
\midrule
\multicolumn{2}{@{}l}{Initial memory ($p_x{=}128$, $k{=}16$)} & \multicolumn{2}{r@{}}{13.1\,MB} \\
\multicolumn{2}{@{}l}{Full fused KV (C2C, 2K input)} & \multicolumn{2}{r@{}}{${\sim}268$\,MB} \\
\bottomrule
\end{tabular}

\end{table}

\subsection{Per-Request Cost Formulas}
\label{app:cost-formulas}

Let $P(\cdot)$ and $D(\cdot)$ be measured prefill and decode throughputs, $D_b(\student)$ the student's decode throughput with the bridge attached, and $c_M, c_S$ the rental rates of the mentor's and student's GPUs. With $n_{\mathrm{sync}} = \lfloor (L_{\mathrm{out}}-1)/R \rfloor$ (zero when $L_{\mathrm{out}} < R$: short outputs never trigger) and $\tau_{\mathrm{sync}}(R) = \Bytes(R)/\mathrm{bandwidth} + \mathrm{RTT}$,
\begin{align}
\mathrm{Cost}_{\MPname}(R) ={}& c_M\frac{L_{\mathrm{in}}{+}L_{\mathrm{out}}}{P(\mentor)} \nonumber\\
&{}+ c_S\Big[\frac{L_{\mathrm{in}}}{P(\student)}+\frac{L_{\mathrm{out}}}{D_b(\student)}+n_{\mathrm{sync}}\,\tau_{\mathrm{sync}}(R)\Big],\label{eq:costmp}\\
\mathrm{Cost}_{\mathrm{T2T}} ={}& c_M\Big[\frac{L_{\mathrm{in}}}{P(\mentor)}+\frac{L_g}{D(\mentor)}\Big] \nonumber\\
&{}+ c_S\Big[\frac{L_{\mathrm{in}}{+}L_g}{P(\student)}+\frac{L_{\mathrm{out}}}{D(\student)}\Big].\label{eq:costt2t}
\end{align}
Two structural facts follow. Incremental prefill touches each generated token exactly once, batched into one parallel forward per refresh, so the mentor's total compute is independent of $R$; only the synchronization term grows as $R$ shrinks. And T2T pays $L_g/D(\mentor)$, mentor decoding, the most expensive per-token operation in the pipeline, which \method{} never performs. rT2T (replace-style) turns that one-time decoding tax into a per-refresh tax:
\begin{align}
\mathrm{Cost}_{\mathrm{rT2T}}(R) ={}& \mathrm{Cost}_{\mathrm{T2T}}
 + c_M\Big[\frac{L_{\mathrm{out}}}{P(\mentor)}+n_{\mathrm{sync}}\frac{L_g'}{D(\mentor)}\Big] \nonumber\\
&{}+ c_S\Big[\sum_{j=1}^{n_{\mathrm{sync}}}\frac{L_g'+L_{\mathrm{in}}+jR}{P(\student)} \nonumber\\
&\qquad{}+ n_{\mathrm{sync}}\Big(\frac{L_g'}{D(\mentor)}+\mathrm{RTT}\Big)\Big],\label{eq:costrt2t}
\end{align}
where the sum is the replace-style full re-prefill and the second $c_S$ term is the student idling while the mentor decodes each hint; the accumulation variant replaces the sum by $n_{\mathrm{sync}} L_g'/P(\student)$. The mentor decoding term dominates ($>$80\% of total), which is why the replace-vs-accumulate cost difference is second order (Appendix~\ref{app:rt2t}).

\subsection{Measured Throughput and Prices}
\label{app:throughput}

All throughputs are measured with vLLM 0.19.1 in bf16 at batch size 1 and 8K-token context on the deployment stack of Section~\ref{sec:setup-exp}; rental rates are public on-demand quotes with their retrieval date. Costs are reported after converting to \$/Mtok (dollars per million output tokens) using each task's measured $L_{\mathrm{out}}$; the conversion never changes a within-task dominance verdict.

\begin{table}[!htbp]
\centering
\caption{Measured serving throughput (vLLM 0.19.1, bf16, batch size 1, 8K context). $D_b$ is decoding with the bridge attached.}
\label{tab:app-throughput}
\footnotesize
\setlength{\tabcolsep}{3pt}
\begin{tabular}{@{}ll rrr@{}}
\toprule
Model & Hardware & $P$ & $D$ & $D_b$ \\
\midrule
Qwen3.5-27B (mentor) & H100 80GB & 8,600 & 54 & --- \\
Qwen3.5-4B (student) & RTX 4090 24GB & 23,500 & 121 & 104 \\
\bottomrule
\end{tabular}

\end{table}

\begin{table}[!htbp]
\centering
\caption{GPU rental prices used in the cost model.}
\label{tab:app-prices}
\small
\setlength{\tabcolsep}{4pt}
\begin{tabular}{@{}l rll@{}}
\toprule
Hardware & \$/h & Source & Quoted \\
\midrule
H100 80GB (mentor) & 2.79 & Lambda on-demand & 2026-07-10 \\
RTX 4090 (student) & 0.44 & RunPod community & 2026-07-10 \\
\bottomrule
\end{tabular}

\end{table}

\subsection{Synchronization Parameters}
\label{app:sync}

The main accounting uses a cross-site leased line, matching the deployment narrative of a data-center mentor and an edge student: 150\,Mbps effective bandwidth, 18\,ms round-trip latency, blocking synchronization, and $k{=}16$ transmitted layers (consistent with ablation row 11). One measured refresh at $R{=}16$ takes 165\,ms end to end. The measurement is the student-observed blocking window and covers every synchronization step: shipping the $R$ new tokens to the mentor, the mentor's incremental prefill of those tokens (one batched parallel forward, not decoding), slot construction, and the slot transfer with its round trip. Because a layer's slots are serialized out as soon as that layer's forward and projection complete, the forward and construction overlap the 142\,ms transfer, and the transfer-dominated formula estimate of 160\,ms (142\,ms transfer $+$ 18\,ms RTT) lands within 3\% of the measured window; the priced tables use the measured 165\,ms as $\tau_{\mathrm{sync}}$. Since $\Bytes(R)$ is tail-dominated, the per-refresh time is nearly constant across $R \in [1, 64]$. This is also why Eq.~\ref{eq:costrt2t} carries an explicit idle term while Eq.~\ref{eq:costmp} does not: at a refresh \MPname{}'s mentor runs one batched forward over $R$ tokens inside the 165\,ms window, whereas rT2T's mentor must decode $L_g' \in \{64, 128\}$ tokens sequentially --- 64--128 dependent forwards --- blocking the student for 1.20--2.39\,s (Appendix~\ref{app:cost-support}); the asymmetry is in the mentor-side operation, not in the accounting. The mentor is billed by \emph{active compute seconds} (a shared serving pool; the mentor is not resident per request); the resident-billing alternative is in Appendix~\ref{app:sensitivity}.

\subsection{Cost Support Numbers}
\label{app:cost-support}

Table~\ref{tab:app-costs} gives the per-request costs behind the dominance verdicts of Figure~\ref{fig:cost} (blocking accounting; quality inputs from Tables~\ref{tab:app-main} and~\ref{tab:app-rsweep}, throughputs and prices from Tables~\ref{tab:app-throughput} and~\ref{tab:app-prices}, lengths from Table~\ref{tab:app-lengths}). The left-end divergence on the long-output rows comes entirely from the synchronization term ($n_{\mathrm{sync}} \propto 1/R$). Dominance is judged on strict numeric comparison per Section~\ref{sec:cost}; rT2T enters no dominance panel because it is cost-dominated at every $R$ (each refresh pays a mentor decoding tax), so Figure~\ref{fig:cost} keeps its two-method verdict clean and rT2T is cited from Table~\ref{tab:app-rt2t-cost} instead. At $R{=}16$ with the quality-best $L_g'{=}128$, rT2T costs 17--28$\times$ T2T and 41--54$\times$ \mpat{16} on the two long-output representatives; a single refresh blocks the student for 1.20\,s ($L_g'{=}64$) or 2.39\,s (128) of mentor decoding versus \MPname{}'s 0.165\,s, and a GovReport request accumulates 52.9--105.1\,s of blocking versus 7.3\,s for \mpat{16}. The structural contrast is worth stating: rT2T's left-end divergence comes from the per-refresh decoding tax $n_{\mathrm{sync}} L_g'/D(\mentor)$, \MPname{}'s from the transfer term --- text refresh is expensive because the mentor must \emph{speak} at every refresh, \MPname{} only because it ships bytes more often.

\begin{table}[!htbp]
\centering
\caption{Per-request cost ($10^{-3}$ dollars, blocking accounting) for the four representative tasks of Figure~\ref{fig:cost}.}
\label{tab:app-costs}
\scriptsize
\setlength{\tabcolsep}{2pt}
\begin{tabular}{@{}l rrrrrrrrr@{}}
\toprule
Task & T2T & \mpat{1} & \mpat{2} & \mpat{4} & \mpat{8} & \mpat{16} & \mpat{32} & \mpat{64} & \mpat{$\infty$} \\
\midrule
A & 1.91 & 0.15 & 0.09 & 0.07 & 0.05 & 0.05 & 0.05 & 0.05 & 0.05 \\
J & 3.44 & 0.59 & 0.53 & 0.51 & 0.49 & 0.49 & 0.49 & 0.49 & 0.49 \\
H & 4.55 & 20.14 & 10.66 & 5.92 & 3.56 & 2.37 & 1.79 & 1.49 & 1.20 \\
K & 5.89 & 15.81 & 8.72 & 5.17 & 3.39 & 2.50 & 2.06 & 1.84 & 1.62 \\
\bottomrule
\end{tabular}

\end{table}

\begin{table*}[!tbp]
\centering
\caption{rT2T per-request cost ($10^{-3}$ dollars, replace-style, blocking; analytically computed from Eq.~\ref{eq:costrt2t} and the measured inputs). Bold marks each task's quality-best configuration (Table~\ref{tab:app-rt2t}). At $\infty$, $n_{\mathrm{sync}}{=}0$ and rT2T equals T2T identically, as on the short-answer tasks (which therefore carry no rows).}
\label{tab:app-rt2t-cost}
\footnotesize
\setlength{\tabcolsep}{4pt}
\begin{tabular}{@{}lc rrrrr c rr@{}}
\toprule
Task & $L_g'$ & $R{=}4$ & $R{=}8$ & $R{=}16$ & $R{=}32$ & $R{=}64$ & $\infty$ & T2T & \mpat{16} \\
\midrule
H WritingBench & 64 & 254.8 & 129.7 & 66.7 & 35.6 & 19.6 & $=$T2T & 4.55 & 2.37 \\
 & 128 & 503.7 & 254.2 & \textbf{128.3} & 66.5 & 34.5 & $=$T2T & 4.55 & 2.37 \\
K GovReport & 64 & 200.8 & 103.4 & 54.7 & 30.3 & 18.1 & $=$T2T & 5.89 & 2.50 \\
 & 128 & 388.0 & 197.0 & \textbf{101.5} & 53.7 & 29.8 & $=$T2T & 5.89 & 2.50 \\
\bottomrule
\end{tabular}

\end{table*}

\subsection{Sensitivity Analyses}
\label{app:sensitivity}

Three accounting variations. \emph{Network:} on a same-site gigabit LAN the synchronization term becomes negligible and the purple segments of Figure~\ref{fig:cost} extend left to $R \approx 2$ (analytic extrapolation of the cost side). \emph{Synchronization mode:} asynchronous prefetch overlaps transfer with decoding and lowers the synchronization term further; the main figure conservatively uses blocking. \emph{Mentor billing:} billing a resident, dedicated mentor for the full request duration instead of active compute seconds reverses the direction of the cost conclusion --- a resident mentor idles between refreshes, and that idle time dwarfs the compute it sells; we report this reversal as is. The cost structures of C2C and LoRA are different in kind rather than in degree and are therefore discussed, not priced, in Figure~\ref{fig:cost}: C2C ships a full fused KV cache (${\sim}268$\,MB at a 2K input, growing with length) and presumes same-machine or same-rack co-location, while LoRA involves no online mentor at all --- its serving cost is a strict lower bound among the compared methods, and Table~\ref{tab:generality} shows what that saving costs in quality.

% ======================================================================
\section{Additional Results and Case Studies}
\label{app:extra-results}

\subsection{Per-Dataset Generality}
\label{app:gen-detail}

\begin{figure*}[!tbp]
\centering
\includegraphics[width=\textwidth]{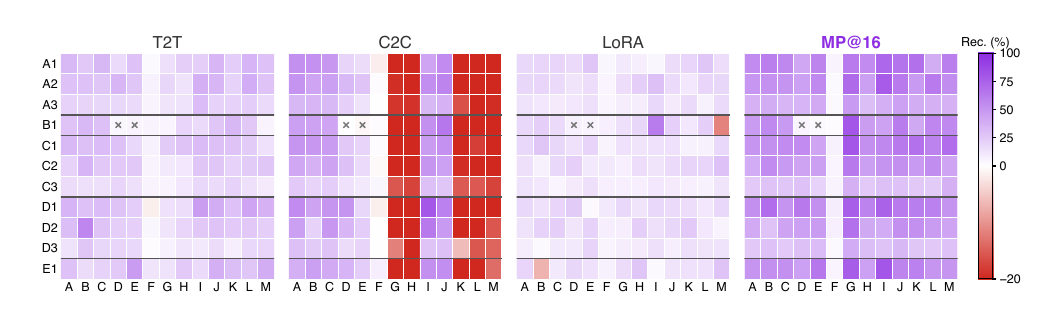}
\caption{Per-dataset recovery for all eleven pairs (rows, grouped by mentor group) on all thirteen datasets (columns) under the four methods; color is recovery clipped to $[-20, 100]$ as in Figure~\ref{fig:quadrant}, red marking harm. $\times$ marks the two capability-limited B1 cells registered in Appendix~\ref{app:pairs}. Numbers behind every cell are in Tables~\ref{tab:app-main} and~\ref{tab:app-pairs-1}--\ref{tab:app-pairs-5}.}
\label{fig:gen}
\end{figure*}

Figure~\ref{fig:gen} expands Table~\ref{tab:generality} to the per-dataset level: eleven pairs $\times$ thirteen datasets for each of the four methods, on the recovery color scale of Figure~\ref{fig:quadrant}. Three patterns repeat across every mentor group. The C2C panel shows the sign-flip columns (G, H, K, L, M) in red for all eleven pairs --- the harm of static latent guidance on long outputs is not a property of one pair. The \MPname{} panel keeps those columns purple everywhere, with saturation fading within each mentor group as the capacity gap grows (the fit-interval pattern of Section~\ref{sec:analysis}). And the F column stays near white in all four panels: no interface rescues a capability the student lacks.

\subsection{Probe Curves}
\label{app:probe-fig}

Figure~\ref{fig:app-probe} plots the registered probe values of Table~\ref{tab:app-probe} (protocol in Appendix~\ref{app:probe}): the stale curve decays from 0.79 to 0.47 (depth 0.32) while the fresh curve holds 0.76--0.80, and the widening wedge between them is $\StaleGap(t)$ --- the guidance drifts away from the task it is supposed to describe even though the input never changes.

\begin{figure}[!htbp]
\centering
\includegraphics[width=\columnwidth]{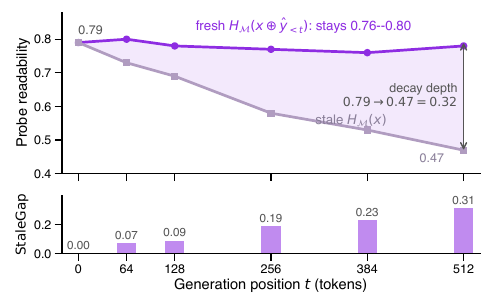}
\caption{StaleGap probe on D-IF (enlarged version of Figure~\ref{fig:oracle}b; linear readout on frozen mentor states, independent of the bridge). Top: fresh and stale probe readability over generation position; the shaded wedge is $\StaleGap(t)$ and the bracket marks the 0.32 decay depth. Bottom: $\StaleGap(t)$ values (fresh $-$ stale, Table~\ref{tab:app-probe}).}
\label{fig:app-probe}
\end{figure}

\subsection{Learned Staleness Trigger (Stage-2 Outlook)}
\label{app:trigger}

\begin{figure}[!htbp]
\centering
\includegraphics[width=\columnwidth]{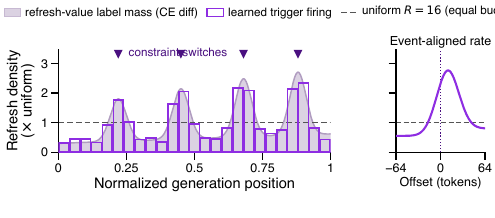}
\caption{Stage-2 outlook (illustrative pilot; no quantitative claim). Background: density of offline refresh-value labels (CE difference between decoding under stale and fresh memory). Bars: where the learned trigger fires at the $R{=}16$ equal budget; the dashed line is the uniform schedule. Firing concentrates at constraint switches; the inset shows the event-aligned rate on the $R{=}16$ window scale.}
\label{fig:app-trigger}
\end{figure}

All experiments in this paper use a fixed refresh interval; a learned trigger (\MPname{}-ST) that decides \emph{when} to refresh is future work, and we register here a preliminary, prospective observation from a stage-2 pilot. The trigger is trained on an offline refresh-value target: for every window, the cross-entropy difference between decoding under the stale and the fresh memory version in the versioned store. A lightweight predictor reads bridge statistics, including attention on recent tail slots, and its threshold is calibrated to match the refresh count of the fixed schedule, so the comparison is at equal budget. Figure~\ref{fig:app-trigger} shows the qualitative behavior: firing concentrates at constraint switches and topic boundaries rather than uniformly, suggesting that a content-aware schedule could buy back part of the small-$R$ quality at fixed cost. No collection table registers \MPname{}-ST numbers yet, and the figure makes no quantitative claim.

\subsection{Qualitative Cases}
\label{app:cases}

\begin{figure*}[!tbp]
\centering
\includegraphics[width=\textwidth]{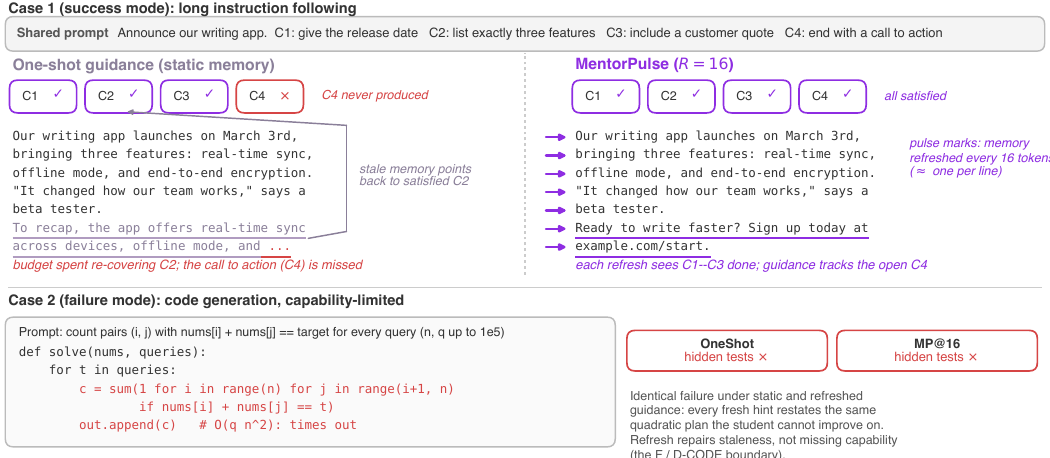}
\caption{Two illustrative cases (more in the full logs, to be released). Top: on long instruction following, static guidance points the student back to a satisfied constraint (gray underline and arrow) and a later constraint is missed (red); with $R{=}16$ the refreshed memory reflects the text written so far and the remaining constraint is completed (purple). Bottom: on a capability-limited coding task both conditions produce the same flawed algorithm --- fresh guidance does not create missing capability.}
\label{fig:app-cases}
\end{figure*}

Figure~\ref{fig:app-cases} shows two representative generation pairs, one from each side of the applicability boundary. In the success case (long instruction following), the student under \oneshot{} begins correctly, then follows the stale memory back to an already satisfied constraint and never produces a later one; under \mpat{16} each refresh sees which constraints are done and the guidance tracks the ones still open. In the failure case (code generation, the capability-limited mode), \oneshot{} and \mpat{16} fail identically: every fresh hint restates a plan the student cannot execute better, so refresh repairs nothing --- refresh fixes staleness, not missing capability, matching the F and D-CODE boundaries in the main text.

\end{document}